\documentclass[11pt,a4paper]{article}
\usepackage{times,latexsym}
\usepackage{url}
\usepackage[T1]{fontenc}

\usepackage[acceptedWithA]{tacl2018v2}

\usepackage{enumitem}
\usepackage{microtype}
\usepackage{footnote}

\usepackage[font=small]{caption}
\usepackage[labelformat=simple]{subcaption}

\usepackage{multirow}
\usepackage{adjustbox}
\usepackage{booktabs}
\usepackage{graphicx}
\usepackage{cleveref}

\usepackage{textcomp}
\usepackage{siunitx}

\makesavenoteenv{tabular}
\makesavenoteenv{table}

\newcommand{\comment}[1]{}

\newcommand{\minus}{\scalebox{0.7}[1.0]{$-$}}

\newcommand{\bettertilde}{\raise.17ex\hbox{$\scriptstyle\sim$}}

\title{ByT5: Towards a Token-Free Future with Pre-trained \\ Byte-to-Byte Models}

\author{
 Linting Xue\Thanks{Equal contribution.} , Aditya Barua\footnotemark[1] , Noah Constant\footnotemark[1] , Rami Al-Rfou\footnotemark[1] , \\ {\bf Sharan Narang, Mihir Kale, Adam Roberts, Colin Raffel} \\
 Google Research \\
 {\texttt \{lintingx, adityabarua, nconstant, rmyeid, sharannarang, mihirkale, adarob\}} \\
 {\texttt @google.com, craffel@gmail.com}
}

\date{}

\begin{document}
\maketitle
\begin{abstract}
Most widely-used pre-trained language models operate on sequences of tokens corresponding to word or subword units.
By comparison, \textit{token-free} models that operate directly on raw text (bytes or characters) have many benefits: they can process text in any language out of the box, they are more robust to noise, and they minimize technical debt by removing complex and error-prone text preprocessing pipelines.
Since byte or character sequences are longer than token sequences, past work on token-free models has often introduced new model architectures designed to amortize the cost of operating directly on raw text.
In this paper, we show that a standard Transformer architecture can be used with minimal modifications to process byte sequences.
We characterize the trade-offs in terms of parameter count, training FLOPs, and inference speed, and show that byte-level models are competitive with their token-level counterparts.
We also demonstrate that byte-level models are significantly more robust to noise and perform better on tasks that are sensitive to spelling and pronunciation.
As part of our contribution, we release a new set of pre-trained byte-level Transformer models based on the T5 architecture, as well as all code and data used in our experiments.\footnote{\label{fn:code}\url{https://github.com/google-research/byt5}}
\end{abstract}

\section{Introduction}

\begin{figure*}[h!]
\centering
\includegraphics[width=0.9\textwidth, trim=0 3 0 4, clip]{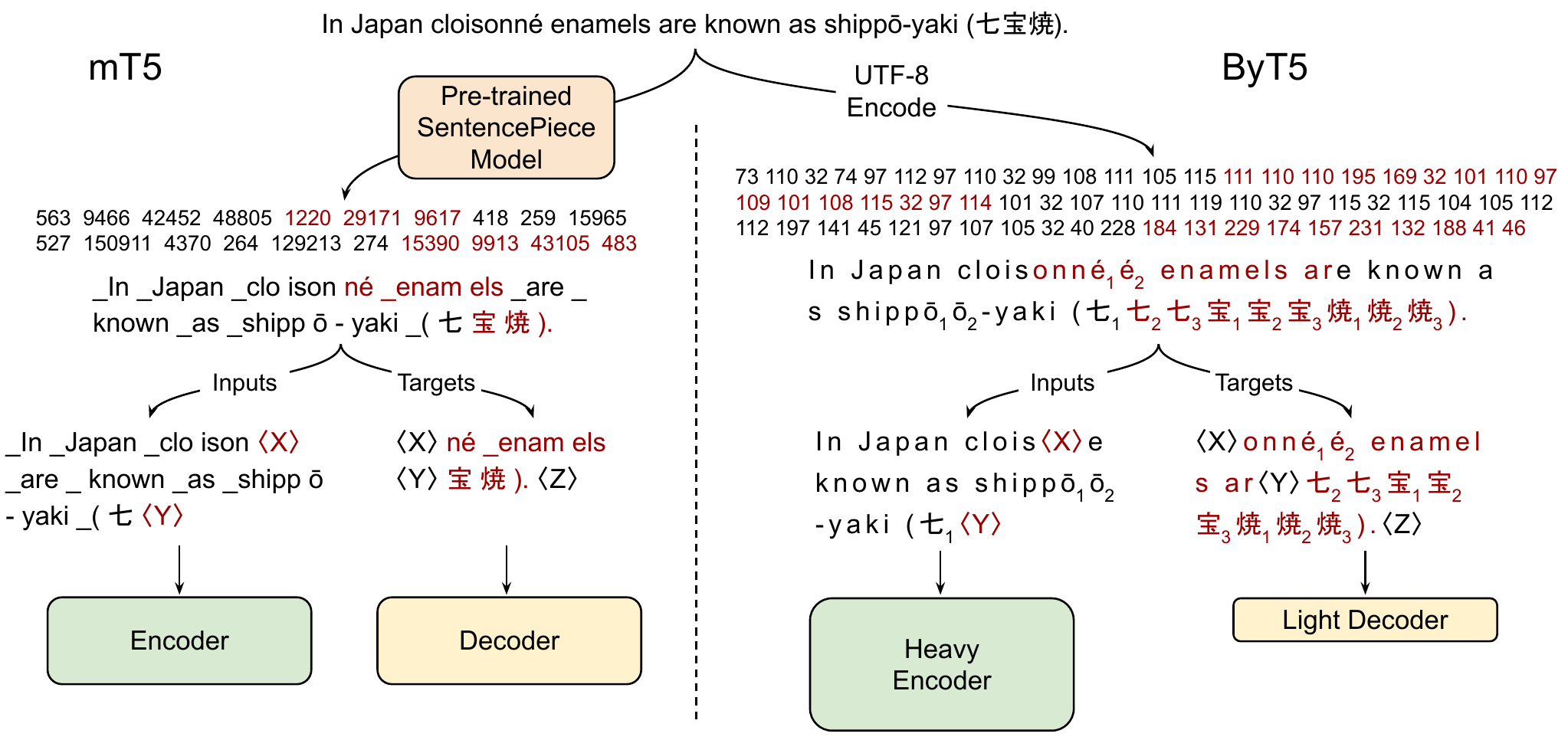}
     \caption{Pre-training example creation and network architecture of mT5 \cite{mt5} vs.~ByT5 (this work). \textbf{mT5}: Text is split into SentencePiece tokens, spans of \bettertilde{}$3$ tokens are masked (red), and the encoder/decoder transformer stacks have equal depth. \textbf{ByT5}: Text is processed as UTF-8 bytes, spans of \bettertilde{}$20$ bytes are masked, and the encoder is $3\times$ deeper than the decoder. $\langle$X$\rangle$, $\langle$Y$\rangle$, and $\langle$Z$\rangle$ represent sentinel tokens.}
     \label{fig:text-processing}
\end{figure*}

An important consideration when designing NLP models is the way that text is represented.
One common choice is to assign a unique \textit{token} ID to each word in a fixed finite vocabulary.
A given piece of text is thus converted into a sequence of tokens by a \textit{tokenizer} before being fed into a model for processing.
An issue with using a fixed vocabulary of words is that there is no obvious way to process a piece of text that contains an \textit{out-of-vocabulary} word.
A standard approach is to map all unknown words to the same \texttt{<UNK>} token, but this prevents the model from distinguishing between different out-of-vocabulary words.

Subword tokenizers \cite{sennrich-etal-2016-neural, wu-2016-gnmt, kudo-richardson-2018-sentencepiece} present an elegant solution to the out-of-vocabulary problem.
Instead of mapping each word to a single token, subword tokenizers decompose words into smaller subword units with a goal of minimizing the total length of the token sequences for a fixed vocabulary size.
As an example, a subword tokenizer might tokenize the word \textit{doghouse} as the pair of tokens \textit{dog} and \textit{house} even if \textit{doghouse} is not in the subword vocabulary.
This flexibility has caused subword tokenizers to become the \textit{de facto} way to tokenize text over the past few years.

However, subword tokenizers still exhibit various undesirable behaviors.
Typos, variants in spelling and capitalization, and morphological changes can all cause the token representation of a word or phrase to change completely, which can result in mispredictions.
Furthermore, unknown characters (e.g.\ from a language that was not used when the subword vocabulary was built) are typically out-of-vocabulary for a subword model.

A more natural solution that avoids the aforementioned pitfalls would be to create \textit{token-free} NLP models that do not rely on a learned vocabulary to map words or subword units to tokens.
Such models operate on raw text directly.
We are not the first to make the case for token-free models, and a more comprehensive treatment of their various benefits can be found in recent work by \citet{clark2021canine}.
In this work, we make use of the fact that text data is generally stored as a sequence of bytes.
Thus, feeding byte sequences directly into the model enables the processing of arbitrary text sequences.
This approach is well-aligned with the philosophy of end-to-end learning, which endeavors to train models to directly map from raw data to predictions.
It also has a concrete benefit in terms of model size: the large vocabularies of word- or subword-level models often result in many parameters being devoted to the vocabulary matrix.
In contrast, a byte-level model by definition only requires $256$ embeddings.
Migrating word representations out of a sparse vocabulary matrix and into dense network layers should allow models to generalize more effectively across related terms (e.g.~\textit{book} / \textit{books}) and orthographic variations.
Finally, from a practical standpoint, models with a fixed vocabulary can complicate adaptation to new languages and new terminology, whereas, by definition, token-free models can process any text sequence.

The main drawback of byte-level models is that byte sequences tend to be significantly longer than token sequences. Since computational costs of machine learning models tend to scale with sequence length, much previous work on character- and byte-level models has explored ways to process long sequences efficiently using convolutions with pooling \cite{xiang-2015-character, lee-etal-2017-fully} or adaptive computation time \cite{graves-2016-adaptive}.

In this work, we take a simpler approach and show that the Transformer architecture can be straightforwardly adapted to process byte sequences without a dramatically unfavorable increase in computational cost.
We focus on the T5 framework \cite{raffel-2020-t5}, where all text-based NLP problems are cast to a text-to-text format.
This approach makes it simple to tackle an NLP task by generating a sequence of bytes conditioned on some input bytes.
Our proposed \textbf{ByT5} architecture is described in \cref{sec:byt5}. The design stays fairly close to mT5 (the multilingual variant of T5 introduced by \citet{mt5}), with the differences illustrated in \cref{fig:text-processing}.
Through extensive experiments on a diverse set of English and multilingual tasks (presented in \cref{sec:core_results}), we show that ByT5 is competitive with a subword-level baseline, despite being pre-trained on $4\times$ less text.
We also confirm in \cref{sec:noise} that byte-level models are more robust to corruptions of the input text.
Throughout, we characterize the trade-offs of our design decisions in terms of computational cost and parameter count, discussed in more detail in \cref{sec:ablations,sec:speed}.
The end result is a set of pre-trained ByT5 models that we release alongside this paper.

\section{Related Work}
\label{sec:related-work}

The early neural language models of \citet{sutskever-2011-generating} and \citet{graves-2013-generating} operated directly on character sequences. This precedent led many to use character-level language modeling as a benchmark to evaluate neural architectures \cite{kalchbrenner-2016-bytenet, chung-2017-hierarchical, ha-2017-hypernets, zilly-2017-recurrent-highway-nets, melis-2018-sota-eval, al-rfou-2019-character-lm}. \citet{Choe2019BridgingTG} showed byte language models can match the perplexity of word-level models when given the same parameter budget. However, standard practice in real-world scenarios has remained to use word- or subword-level models.

A number of \textit{character-aware} architectures have been proposed that make use of character-level features but still rely on a tokenizer to identify word boundaries. These approaches include ELMo \cite{peters-etal-2018-deep}, CharacterBERT \cite{el-boukkouri-etal-2020-characterbert} and many others \cite{ling-2015-character, chung-etal-2016-character, kim-2016-character-aware, rafal-2016-exploring, wang-2020-nmt-with-byte, wei-2021-training}.
Separately, some work has endeavored to ameliorate issues with tokenization, for example by adapting vocabularies to new languages \cite{garcia-2021-continual} or randomly choosing different subword segmentations to improve robustness in low-resource and out-of-domain settings \cite{kudo-2018-subword}.
These methods do not meet our goal of simplifying the NLP pipeline by removing text preprocessing.

There have been a few recent efforts to develop general-purpose token-free pre-trained language models for transfer learning.\footnote{Previous work has also developed token-free approaches for specific tasks: \citet{gillick-etal-2016-multilingual} for span labeling, \citet{li-2019-bytes} for speech recognition and synthesis, and many authors for machine translation \cite{lee-etal-2017-fully, costa-jussa-etal-2017-byte, cherry-etal-2018-revisiting, shaham-levy-2021-neural}.}
\citet{akbik-etal-2018-contextual} show strong results on sequence labeling with character-level pre-training and release models covering four languages. More recently, \citet{clark2021canine} develop \textsc{Canine}, which shows gains over multilingual BERT by working with characters instead of word-piece tokens, though the ``\textsc{Canine}-S'' model still uses a tokenizer during pre-training to define targets for the masked language modeling task. Our work differs from these in that (i)~we train encoder-decoder models that extend to generative tasks, (ii)~our models work directly with UTF-8 bytes, and (iii)~we explore the effect of model scale, training models beyond $10$ billion parameters.

\section{ByT5 Design}
\label{sec:byt5}

Our goal in designing ByT5 is to take an existing token-based model and perform the minimal set of modifications to make it token-free, thereby limiting experimental confounds.
We base ByT5 on the recent mT5 model \cite{mt5}, which was trained on mC4 (a large corpus of unlabeled multilingual text data) and achieved state-of-the-art on many community benchmarks.
We release ByT5 in five sizes analogous to T5 and mT5 (Small, Base, Large, XL, XXL).
We aim for ByT5 to cover the same use cases as mT5: it is a general-purpose pre-trained text-to-text model covering 100+ languages.
We expect ByT5 will be particular useful for tasks operating on short-to-medium length text sequences (a few sentences or less), as these will incur less slowdown in fine-tuning and inference.

\subsection{Changes from mT5}

Compared to mT5, we make the following key changes in designing ByT5. First and foremost, we dispense with the SentencePiece \cite{kudo-richardson-2018-sentencepiece} vocabulary and feed \mbox{UTF-8} bytes directly into the model without any text preprocessing. The bytes are embedded to the model hidden size using a vocabulary of $256$ possible byte values. An additional $3$ IDs are reserved for special tokens: padding, end-of-sentence, and an unused \texttt{<UNK>} token that we include only for convention.

Second, we modify the pre-training task.
mT5 uses the ``span corruption'' pre-training objective first proposed by \citet{raffel-2020-t5} where spans of tokens in unlabeled text data are replaced with a single ``sentinel'' ID and the model must fill in the missing spans.
Rather than adding 100 new tokens for the sentinels, we find it sufficient to reuse the final 100 byte IDs.
While mT5 uses an average span length of $3$ subword tokens, we find that masking longer byte-spans is valuable.
Specifically, we set our mean mask span length to $20$ bytes, and show ablations of this value in \cref{sec:ablations}.

Third, we find that ByT5 performs best when we decouple the depth of the encoder and decoder stacks. While T5 and mT5 used ``balanced'' architectures, we find byte-level models benefit significantly from a ``heavier'' encoder. Specifically, we set our encoder depth to $3$ times that of the decoder. Intuitively, this heavier encoder makes the model more similar to encoder-only models like BERT\@. By decreasing decoder capacity, one might expect quality to deteriorate on tasks like summarization that require generation of fluent text. However, we find this is not the case, with heavy-encoder byte models performing better on both classification and generation tasks. We ablate the effect of encoder/decoder balance in \cref{sec:ablations}.

As not all byte sequences are legal according to the \mbox{UTF-8} standard, we drop any illegal bytes in the model's output\footnote{This is achieved with the Python bytes-decoding function \texttt{bytes.decode("utf-8", errors="ignore")}.} (though we never observed our models predicting illegal byte sequences in practice). Apart from the above changes, we follow mT5 in all settings. Like mT5, we set our sequence length to $1024$ (bytes rather than tokens), and train for $1$ million steps over batches of $2^{20}$ tokens.

\subsection{Comparing the Models}

Our goal in this paper is to show that straightforward modifications to the Transformer architecture can allow for byte-level processing while incurring reasonable trade-offs in terms of cost.
Characterizing these trade-offs requires a clear definition of what is meant by ``cost'', since there are many axes along which it is possible to measure a model's size and computational requirements.

\begin{table}[t!]
\hspace{-8pt}
\centering
\resizebox{1.015\columnwidth}{!}{
\footnotesize
\setlength\tabcolsep{1pt}
\begin{tabular}{lccccccc}
\toprule
& & \multicolumn{3}{c}{\textbf{mT5}} & \multicolumn{3}{c}{\textbf{ByT5}} \\
\cmidrule(l{4pt}r{4pt}){3-5} \cmidrule(l{4pt}r{4pt}){6-8}
Size & Param & Vocab & d\textsubscript{model}\,/\,d\textsubscript{ff} & Enc/Dec & Vocab & d\textsubscript{model}\,/\,d\textsubscript{ff} & Enc/Dec \\
\midrule
Small & 300M  & 85\%    & 512\,/\,1024    & 8/8  & 0.3\% & 1472\,/\,3584  & 12/4    \\
Base  & 582M  & 66\%  & 768\,/\,2048  & 12/12 & 0.1\% & 1536\,/\,3968  & 18/6    \\
Large & 1.23B & 42\% & 1024\,/\,2816 & 24/24 & 0.06\% & 1536\,/\,3840  & 36/12   \\
XL    & 3.74B & 27\%   & 2048\,/\,5120  & 24/24 & 0.04\% & 2560\,/\,6720  & 36/12   \\
XXL   & 12.9B & 16\%  & 4096\,/\,10240  & 24/24 & 0.02\% & 4672\,/\,12352 & 36/12  \\
\bottomrule
\end{tabular}}
\caption{Comparison of mT5 and ByT5 architectures. For a given named size (e.g.~``Large''), the total numbers of parameters and layers are fixed. ``Vocab'' shows the percentage of vocabulary-related parameters, counting both the input embedding matrix and the decoder softmax layer. ByT5 moves these parameters out of the vocabulary and into the transformer layers, as well as shifting to a $3$:$1$ ratio of encoder to decoder layers.}
\label{tab:model_params}
\end{table}

Models that use a word or subword vocabulary typically include a vocabulary matrix that stores a vector representation of each token in the vocabulary.
They also include an analogous matrix in the output softmax layer.
For large vocabularies (e.g.\ those in multilingual models), these matrices can make up a substantial proportion of the model's parameters.
For example, the vocabulary and softmax output matrices in the \mbox{mT5-Base} model amount to $256$ million parameters, or about 66\% of the total parameter count.
Switching to a byte-level model allows allocating these parameters elsewhere in the model, e.g.\ by adding layers or making existing layers ``wider''.
To compensate for reduction in total parameter count due to changing from a token-based to token-free model, we adjust our ByT5 model hidden size (d\textsubscript{model}) and feed-forward dimensionality (d\textsubscript{ff}) to be \textit{parameter-matched} with mT5, while maintaining a ratio of roughly $2.5$ between d\textsubscript{ff} and d\textsubscript{model}, as recommended by \citet{kaplan2020scaling}.
\Cref{tab:model_params} shows the resulting model architectures across all five model sizes.

Separately, as previously mentioned, changing from word or subword sequences to byte sequences will increase the (tokenized) sequence length of a given piece of text.
The self-attention mechanism at the core of the ubiquitous Transformer architecture \cite{vaswani2017attention} has a quadratic time and space complexity in the sequence length, so byte sequences can result in a significantly higher computational cost.
While recurrent neural networks and modified attention mechanisms \cite{tay2020efficient} can claim a better computational complexity in the sequence length, the cost nevertheless always scales up as sequences get longer.

Thus far, we have been discussing easy-to-measure quantities like the parameter count and FLOPs.
However, not all FLOPs are equal, and the real-world cost of a particular model will also depend on the hardware it is run on.
One important distinction is to identify operations that can be easily parallelized (e.g.\ the encoder's fully-parallelizable processing) and those that cannot (e.g.\ autoregressive sampling in the decoder during inference).
For byte-level encoder-decoder models, if the decoder is particularly large, autoregressive sampling can become comparatively expensive thanks to the increased length of byte sequences.
Relatedly, mapping an input token to its corresponding vector representation in the vocabulary matrix is essentially ``free'' in terms of FLOPs since it can be implemented by addressing a particular row in memory.
Therefore, reallocating parameters from the vocabulary matrix to the rest of the model will typically result in a model that requires more FLOPs to process a given input sequence (see \cref{sec:speed} for detailed comparison).

\begin{figure*}[h!]
\centering
\includegraphics[width=\textwidth, trim=7 7 7 7, clip]{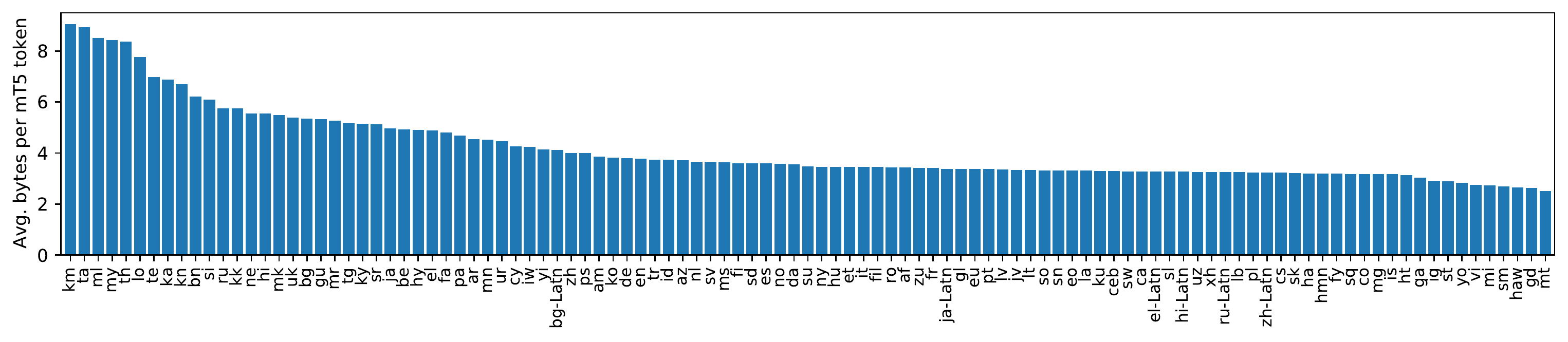}
     \caption{Per-language compression rates of the mT5 SentencePiece vocabulary, measured over the mC4 pre-training corpus. For each language, we measure the ratio of UTF-8 bytes to tokens over all mC4 data in that language.}
     \label{fig:compression-rates}
\end{figure*}

Finally, we note that another important metric is data efficiency, i.e.\ how much data is required for the model to reach a good solution.
For NLP problems, this can be measured either in terms of the number of tokens or the amount of raw text seen during training.
Specifically, a byte-level model trained on the same number of tokens as a word- or subword-level model will have been trained on less text data.
In Figure~\ref{fig:compression-rates}, we show the compression rates of mT5 SentencePiece tokenization, measured as the ratio of UTF-8 bytes to tokens in each language split of the mC4 corpus used in pre-training. This ratio ranges from $2.5$ (Maltese) to $9.0$ (Khmer). When considering the mC4 corpus as a whole, sampled according to the mT5 pre-training mixing ratios, we have an overall compression rate of $4.1$ bytes per SentencePiece token. On the one hand, this $4\times$ lengthening could be seen as an advantage for ByT5: with longer sequences, the model gets more compute to spend encoding a given piece of text. On the other hand, given a fixed input sequence length and number of training steps, the model will be exposed to roughly $4\times$ less actual text during pre-training.

With these factors in mind, we choose to focus on the following measures of efficiency in our experiments: parameter count, inference time, and pre-training efficiency.
Parameter count is a simple and easy-to-measure quantity that directly relates to the amount of memory required to use a model.
Inference time is a real-world measurement of the model's computational cost that represents a ``worst-case'' measurement for byte-level models given the potential additional cost of autoregressive sampling.
Finally, pre-training efficiency allows us to measure whether byte-level models can learn a good solution after seeing less pre-training data.

\section{Core Results}
\label{sec:core_results}

In this section, we compare ByT5 against mT5 on a wide range of tasks. We show that ByT5 is competitive with mT5 on standard English and multilingual NLP benchmarks and outperforms mT5 at small model sizes. Additionally, ByT5 excels on free-form generation tasks and word-level tasks.

For each downstream task, we fine-tune mT5 and ByT5 models for $262{,}144$ steps, using a constant learning rate of $0.001$ and a dropout rate of $0.1$.\footnote{For some tasks we observed clear saturation or overfitting on validation set metrics, and shortened the total fine-tuning steps: $70{,}000$ for Dakshina, $30{,}000$ for TweetQA, and $10{,}000$ for the \textsc{Sigmorphon} tasks.} We use a batch size of $2^{17}$ tokens by default, but increased this to $2^{20}$ for several tasks with larger training sets (GLUE, SuperGLUE, XNLI, TweetQA), and decreased to $2^{16}$ for the Dakshina task. In all cases, we select the best model checkpoint based on validation set performance.

\subsection{English Classification Tasks}

\begin{table}[t!]
\centering
\resizebox{0.7\columnwidth}{!}{
\footnotesize
\begin{tabular}{lcccccc}
\toprule
& \multicolumn{2}{c}{GLUE} & \multicolumn{2}{c}{SuperGLUE} \\
\cmidrule(lr){2-3} \cmidrule(lr){4-5}
Model & mT5 & ByT5 & mT5 & ByT5 \\
\midrule
 Small & 75.6 & \textbf{80.5} & 60.2 & \textbf{67.8} \\
 Base & 83.0 & \textbf{85.3} & 72.5 & \textbf{74.0} \\
 Large & \textbf{87.6} & 87.0 & \textbf{81.9} & 80.4 & \\
 XL & \textbf{88.7} & 87.9 & \textbf{84.7} & 83.2 \\
 XXL & \textbf{90.7} & 90.1 & \textbf{89.2} & 88.6 \\
\bottomrule
\end{tabular}}
\caption{mT5 and ByT5 performance on GLUE and SuperGLUE\@. For each benchmark, we fine-tune a single model on a mixture of all tasks, select the best checkpoint per task based on validation set performance, and report average validation set scores over all tasks.}
\label{tab:english_classification}
\end{table}

On the widely-adopted GLUE \cite{wang2019glue} and SuperGLUE \cite{wang2019superglue} text classification benchmarks, we find ByT5 beats mT5 at the Small and Base sizes, but mT5 has the advantage at larger sizes, as shown in \cref{tab:english_classification}. The strong performance of ByT5 at smaller sizes likely stems from the large increase in dense parameters over mT5. While the overall models are parameter-matched, most of the mT5 Small and Base parameters are ``locked'' in vocab-related matrices and are only accessed when a particular token is present. We suspect that replacing these with ``dense'' parameters activated across all examples encourages more efficient parameter usage and sharing.

\subsection{English Generation Tasks}

We also compare ByT5 with mT5 on three English generative tasks. XSum \cite{narayan-etal-2018-dont} is an abstractive summarization task requiring models to summarize a news article in a single sentence. For better comparison to recent work, we adopt the version of the task defined in the GEM benchmark \cite{gehrmann2021gem}. TweetQA \cite{xiong-etal-2019-tweetqa} is an abstractive question-answering task built from tweets mentioned in news articles. This tests understanding of the ``messy'' and informal language of social media. Finally, DROP \cite{dua-etal-2019-drop} is a challenging reading comprehension task that requires numerical reasoning.

\begin{table}[t!]
\centering
\resizebox{\columnwidth}{!}{
\footnotesize
\begin{tabular}{l cc cc cc}
\toprule
& \multicolumn{2}{c}{GEM-XSum} & \multicolumn{2}{c}{TweetQA} & \multicolumn{2}{c}{DROP} \\
& \multicolumn{2}{c}{(BLEU)} & \multicolumn{2}{c}{(BLEU-1)} & \multicolumn{2}{c}{(F1 / EM)} \\
\cmidrule(lr){2-3} \cmidrule(lr){4-5} \cmidrule(lr){6-7}
Model & mT5 & ByT5 & mT5 & ByT5 & mT5 & ByT5 \\
\midrule
Small & 6.9 & \textbf{9.1} & 54.4 & \textbf{65.7} & 40.0 / 38.4 & \textbf{66.6 / 65.1} \\
Base & 8.4 & \textbf{11.1} & 61.3 & \textbf{68.7} & 47.2 / 45.6 & \textbf{72.6 / 71.2} \\
Large & 10.1 & \textbf{11.5} & 67.9 & \textbf{70.0} & 58.7 / 57.3 & \textbf{74.4 / 73.0} \\
XL & 11.9 & \textbf{12.4} & 68.8 & \textbf{70.6} & 62.7 / 61.1 & \textbf{{68.7 / 67.2}} \\
XXL & 14.3 & \textbf{15.3} & 70.8 & \textbf{72.0} & 71.2 / 69.6 & \textbf{80.0 / 78.5} \\
\bottomrule
\end{tabular}}
\caption{mT5 vs.~ByT5 on three English generation tasks, reporting the best score on the validation set.}
\label{tab:english_generation}
\end{table}

\Cref{tab:english_generation} shows that ByT5 outperforms mT5 on each generative task across all model sizes. On \mbox{GEM-XSum}, ByT5 comes close ($15.3$ vs.~$17.0$) to the best score reported by \citet{gehrmann2021gem}, a PEGASUS model \cite{zhang-2020-pegasus} pre-trained specifically for summarization. On TweetQA, ByT5 outperforms ($72.0$ vs.~$67.3$) the BERT baseline of \citet{xiong-etal-2019-tweetqa}. On DROP, ByT5 comes close (EM $78.5$ vs.~$84.1$) to the best result from \citet{chen2020question}, a QDGAT (RoBERTa) model with a specialized numeric reasoning module.

\subsection{Cross-lingual Benchmarks}

\begin{table*}[t!]
\centering
\resizebox{\textwidth}{!}{
\footnotesize
\begin{tabular}{l cc cc cc cc cc}
\toprule
\multicolumn{1}{l}{\multirow{2}{*}{}}  & \multicolumn{2}{c}{Small} & \multicolumn{2}{c}{Base} & \multicolumn{2}{c}{Large} & \multicolumn{2}{c}{XL} & \multicolumn{2}{c}{XXL}        \\

\cmidrule(lr){2-3} \cmidrule(lr){4-5} \cmidrule(lr){6-7} \cmidrule(lr){8-9} \cmidrule(lr){10-11}

 & mT5    & ByT5    & mT5    & ByT5   & mT5    & ByT5    & mT5   & ByT5  & mT5 & ByT5 \\
\midrule 
\multicolumn{11}{l}{\emph{In-language multitask (models fine-tuned on gold data in all target languages)}} \\
\midrule 
WikiAnn NER & 86.4 & \textbf{90.6} & 88.2 & \textbf{91.6} & 89.7 & \textbf{91.8} & 91.3 & \textbf{92.6} & 92.2 & \textbf{93.7} \\
TyDiQA-GoldP & 75.9 / 64.8 & \textbf{82.6 / 73.6} & 81.7 / 71.2 & \textbf{86.4 / 78.0} & 85.3 / 75.3 & \textbf{87.7 / 79.2} & 87.6 / 78.4 & \textbf{88.0 / 79.3} & 88.7 / 79.5 & \multicolumn{1}{r}{\textbf{89.4 / 81.4}} \\
\midrule 
\multicolumn{11}{l}{\emph{Translate-train (models fine-tuned on English data plus translations in all target languages)}} \\
\midrule 
XNLI & 75.3 & \textbf{76.6} & \textbf{80.5} & 79.9 & \textbf{84.4} & 82.8 & \textbf{85.3} & 85.0 & \textbf{87.1} & 85.7 \\
PAWS-X & 87.7 & \textbf{88.6} & \textbf{90.5} & 89.8 & \textbf{91.3} & 90.6 & \textbf{91.0} & 90.5 & 91.5 & \textbf{91.7} \\
XQuAD & 71.3 / 55.7 & \textbf{74.0 / 59.9} & 77.6 / 62.2 & \textbf{78.5 / 64.6} & 81.3 / 66.5 & \textbf{81.4 / 67.4} & 82.7 / 68.1 & \textbf{83.7 / 69.5} & \textbf{85.2 / 71.3} & 84.1 / 70.2 \\
MLQA &	56.6 / 38.8	 & \textbf{67.5 / 49.9} & 	69.7 / 51.0	 & \textbf{71.9 / 54.1} & 	74.0 / 55.0	 & \textbf{74.4 / 56.1} & 	75.1 / 56.6	 & \textbf{75.9 / 57.7} & 	\textbf{76.9} / 58.3	 & \textbf{76.9 / 58.8}	\\										
TyDiQA-GoldP & 49.8 / 35.6 & \textbf{64.2 / 50.6} & 66.4 / 51.0 & \textbf{75.6 / 61.7} & 75.8 / 60.2 & \textbf{80.1 / 66.4} & 80.1 / 65.0 & \textbf{81.5 / 67.6} & \textbf{83.3} / 69.4 & 83.2 / \textbf{69.6} \\
\midrule 
\multicolumn{11}{l}{\emph{Cross-lingual zero-shot transfer (models fine-tuned on English data only)}} \\
\midrule 
XNLI   & 67.5        & \textbf{69.1}        & \textbf{75.4}                 & \textbf{75.4}        & \textbf{81.1}        & 79.7        & \textbf{82.9}        & 82.2        & \textbf{85.0}        & 83.7          \\
PAWS-X  & 82.4        & \textbf{84.0}          & \textbf{86.4}                 & 86.3                 & \textbf{88.9}        & 87.4        & \textbf{89.6}        & 88.6        & 90.0                 & \textbf{90.1} \\
WikiAnn NER & 50.5        & \textbf{57.6} & 55.7        & \textbf{62.0} & 58.5        & \textbf{62.9} & \textbf{65.5} & 61.6        & \textbf{69.2} & 67.7 \\
\bottomrule
\end{tabular}
}
\caption{ByT5 and mT5 performance on a subset of \textsc{xtreme} tasks. Our evaluation setup follows \citet{mt5}. For QA tasks we report F1 / EM scores.}
\label{tab:xtreme}
\end{table*}

Changes to vocabulary and tokenization are likely to affect different languages in different ways. To test the effects of moving to byte-level modeling on cross-lingual understanding, we compare parameter-matched ByT5 and mT5 models on tasks from the popular \textsc{xtreme} benchmark suite \cite{hu_2020_xtreme}. Specifically we evaluate on the same six tasks as \citet{mt5}. These consist of two classification tasks: XNLI \cite{conneau2018xnli} and \mbox{PAWS-X} \cite{yang-etal-2019-paws}, three extractive QA tasks: XQuAD \cite{artetxe-etal-2020-cross}, MLQA \cite{lewis-etal-2020-mlqa} and TyDiQA \cite{clark-etal-2020-tydi}, and one structured prediction task: WikiAnn NER \cite{pan-etal-2017-cross}.

\Cref{tab:xtreme} shows that ByT5 is quite competitive overall. On the most realistic \textit{in-language} setting, where some gold training data is available in all languages, ByT5 surpasses the previous state-of-art mT5 on all tasks and model sizes. On the \textit{translate-train} setting, ByT5 beats mT5 at smaller sizes, but the results are mixed at larger sizes. We report \textit{zero-shot} results for completeness, but emphasize that this setting is less aligned with practical applications, as machine translation is widely available.\footnote{We ignore zero-shot QA tasks, where text-to-text models are known to exhibit illegal predictions \cite{mt5}.} 

We explore per-language breakdowns on two tasks to see how different languages are affected by the switch to byte-level processing. One might expect languages with rich inflectional morphology (e.g.~Turkish) to benefit most from the move away from a fixed vocabulary. We were also curious to see if any patterns emerged regarding language family (e.g.~Romance vs.~Slavic), written script (e.g.~Latin vs.~non-Latin), character set size, or data availability (high vs.~low resource).

\begin{figure}[t!]
\centering
\includegraphics[scale=0.58]{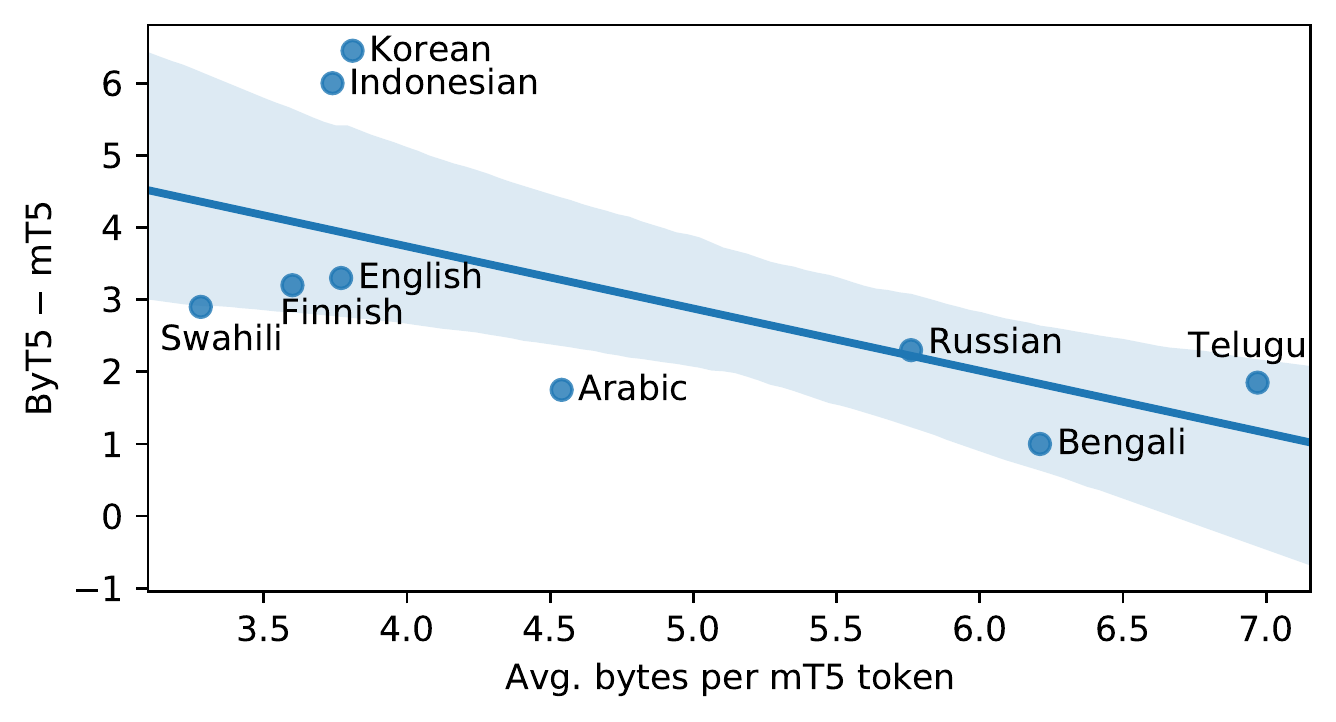}
\includegraphics[scale=0.58]{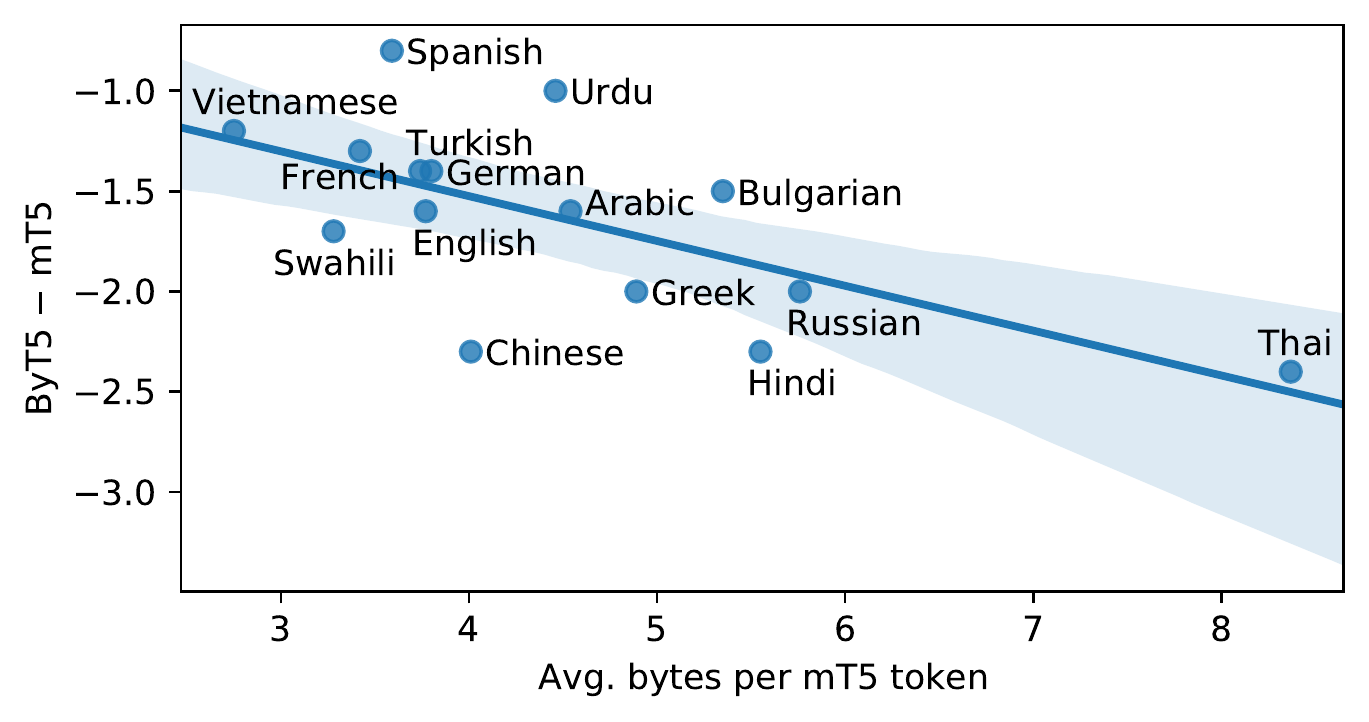}
\caption{Per-language performance gaps between ByT5-Large and mT5-Large, as a function of each language's ``compression rate''. \textbf{Top}: TyDiQA-GoldP gap. \textbf{Bottom}: XNLI zero-shot gap.}
\label{fig:gaps_by_language}
\end{figure}

\Cref{fig:gaps_by_language} shows the per-language gaps between ByT5-Large and mT5-Large on TyDiQA-GoldP and XNLI zero-shot. One notable trend is that the gap is fairly stable across languages. For example, ByT5 is better in each language on TyDiQA-GoldP, while mT5 is consistently better on XNLI\@. Comparing across languages, we observe that languages with a higher SentencePiece token compression rate (e.g.~Thai and Telugu) tend to favor mT5, whereas those with a lower compression rate (e.g.~Indonesian and Vietnamese) tend to favor ByT5. We did not observe any robust trends regarding morphological complexity, language family, script, character set size, or data availability.

\subsection{Word-Level Tasks}

\begin{table}[t!]
\centering
\setlength\tabcolsep{4pt}
\resizebox{1.01\columnwidth}{!}{
\footnotesize
\begin{tabular}{lcccccc}
\toprule
& \multicolumn{2}{c}{Dakshina} & \multicolumn{4}{c}{\textsc{Sigmorphon} 2020} \\
\cmidrule(lr){2-3} \cmidrule(lr){4-7}
& \multicolumn{2}{c}{Transliteration} & \multicolumn{2}{c}{Grapheme-to-Phoneme} & \multicolumn{2}{c}{Inflection} \\
& \multicolumn{2}{c}{CER ($\downarrow$)} & \multicolumn{2}{c}{WER ($\downarrow$) / PER ($\downarrow$)} & \multicolumn{2}{c}{Accuracy ($\uparrow$)} \\
\cmidrule(lr){2-3} \cmidrule(lr){4-5} \cmidrule(lr){6-7}
Model & \hspace{2pt}mT5 & ByT5 & \hspace{8pt}mT5 & ByT5\hspace{8pt} & mT5 & ByT5 \\
\midrule
Small & \hspace{2pt}20.7 & \textbf{9.8} & \hspace{8pt}54.0 / 10.6 & \textbf{14.8} / \textbf{1.8}\hspace{8pt} & 66.5 & \textbf{88.3} \\
Base & \hspace{2pt}19.2 & \textbf{9.9} & \hspace{8pt}46.2 / 7.7 & \textbf{14.0} / \textbf{1.7}\hspace{8pt} & 70.9 & \textbf{89.3} \\
Large & \hspace{2pt}18.1 & \textbf{10.5} & \hspace{8pt}43.5 / 6.7 & \textbf{15.4} / \textbf{1.8}\hspace{8pt} & 75.7 & \textbf{89.7} \\
XL & \hspace{2pt}17.3 & \textbf{10.6} & \hspace{8pt}42.0 / 6.0 & \textbf{14.7} / \textbf{1.8}\hspace{8pt} & 77.4 & \textbf{89.9} \\
XXL & \hspace{2pt}16.6 & \textbf{9.6} & \hspace{8pt}40.1 / 5.4 & \textbf{13.8} / \textbf{1.6}\hspace{8pt} & 78.0 & \textbf{90.9} \\
\bottomrule
\end{tabular}
}
\caption{mT5 vs.~ByT5 on three word-level tasks. Dakshina metrics are reported on the development set to be comparable with \citet{roark-etal-2020-processing}. \textsc{Sigmorphon} metrics are reported on the test sets.}
\label{tab:word_tasks}
\end{table}

Given its direct access to the ``raw'' text signal, we expect ByT5 to be well-suited to tasks that are sensitive to the spelling or pronunciation of text. In this section we test this hypothesis on three word-level benchmarks: (i) transliteration, (ii) grapheme-to-phoneme, and (iii) morphological inflection.

For transliteration, we use the Dakshina benchmark \cite{roark-etal-2020-processing}, which covers $12$ South Asian languages that are traditionally written with Brahmic or Perso-Arabic scripts but may also be written using Latin characters in informal contexts. The single-word transliteration task asks a model to ``translate'' a word from Latin script to native script and measures character error rate. The remaining tasks are \textsc{Sigmorphon} 2020 shared tasks. Multilingual grapheme-to-phoneme conversion \cite{gorman-etal-2020-sigmorphon} covers $15$ languages and requires mapping a word to its pronunciation as phonemes (e.g.~\textit{cat} $\rightarrow$ /k\ae{}t/). Typologically diverse morphological inflection \cite{vylomova-etal-2020-sigmorphon} covers $90$ languages and requires generating a specific inflection of a word (e.g.~\textit{eat} + \textsc{past} $\rightarrow$ \textit{ate}).

We fine-tune mT5 and ByT5 models for each task. For simplicity, we train one multilingual model per task, with a prefix indicating the language in question. \Cref{tab:word_tasks} shows that ByT5 outperforms mT5 by large margins across the board.\footnote{On Dakshina, ByT5 also beats the character-level Transformer baseline of \citet{roark-etal-2020-processing} ($9.6$ vs.~$12.2$). On grapheme-to-phoneme, ByT5 beats the state-of-art model of \citet{yu-etal-2020-ensemble} (PER: $1.6$ vs.~$2.8$). On inflection, ByT5 matches the best single-model \cite{peters-martins-2020-one}.} While it is unsurprising that ``character-aware'' models should excel on tasks around word-internal phenonema, we wish to highlight that these core NLP tasks have often been overlooked in evaluating general-purpose NLP models.

\section{Experiments on Synthetic Noise}
\label{sec:noise}

Text on modern digital platforms is noisy and exhibits complex character-level phenomena such as typos, character repetitions, and non-standard case changes \cite{caswell2020language}. Beyond these, errors can be introduced by NLP systems such as predictive input methods and automatic speech recognition. We have already seen strong ByT5 performance on the ``messy'' text in TweetQA. In this section, we move to even noisier text and explore model performance on inputs that have been corrupted with \emph{artificial} noise of various kinds. Across a range of noising schemes, we find that ByT5 outperforms mT5, demonstrating higher robustness to noise across tasks and languages.

We experiment with five noising schemes: (1)~\textbf{Drop}: Each character (i.e.~Unicode codepoint) has a $10$\% chance of being dropped. (2)~\textbf{Repetitions}: Each character has a 20\% chance of being selected for repetition. If selected, 1--3 repetitions (with equal likelihood) are appended after the original character. (3)~\textbf{Antspeak}: Each character is capitalized and padded with spaces, so ``an~owl'' becomes ``~A~N~~~O~W~L~''. (4)~\textbf{Uppercase}: Each character is converted to uppercase. (5)~\textbf{Random case}: Each character is set to a random case (upper or lower). For the last two noise types, we restrict to languages whose scripts distinguish case.

We first consider the easier setting of \emph{learnable} noise, where noise is applied during both fine-tuning and evaluation. We evaluate on XNLI zero-shot and TyDiQA-GoldP\@. For XNLI, both the premise and hypothesis are noised, and the model predicts an entailment label as usual. For TyDiQA, we add noise to the question and the context, but leave the answer unchanged. Thus, in many cases, the model needs to first locate the noisy answer, and then ``undo'' the noise to produce the target. We fine-tune all models for $30{,}000$ steps following the procedure in \cref{sec:core_results}.

\begin{table}[t!]
\centering
\resizebox{\columnwidth}{!}{
\footnotesize
\setlength\tabcolsep{4pt}
\begin{tabular}{llccc}
\toprule

& & \multicolumn{2}{c}{\textbf{Learnable Noise}} & \textbf{Unseen Noise} \\

\cmidrule(lr){3-4} \cmidrule(lr){5-5}

& & XNLI & TyDiQA- & XNLI \\
& Model & (accuracy) & GoldP (F1) & (accuracy) \\
\midrule

\multirow{2}{*}{Clean}
& mT5 & 81.1 & 85.3 & 81.1 \\
& ByT5 & 79.7 & 87.7 & 79.7 \\

\midrule

\multirow{2}{*}{Drop}
& mT5 & \minus10.2 & \minus24.0 & \minus18.3\\
& ByT5 & \textbf{\minus8.2} & \textbf{\minus19.5} & \textbf{\minus11.4} \\

\midrule

\multirow{2}{*} {Repetitions}
& mT5 & \minus8.5 & \minus9.5 &\minus12.3 \\
& ByT5 & \textbf{\minus4.1} & \textbf{\minus3.0} & \textbf{\minus5.9} \\

\midrule

\multirow{2}{*} {Antspeak}
& mT5 & \minus32.0 & \minus27.7 & \minus34.4 \\
& ByT5 & \textbf{\minus8.7} & \textbf{\minus4.8} & \textbf{\minus24.4}\\

\midrule

\multirow{2}{*} {Uppercase}
& mT5 & \minus7.0 & \minus8.0 & \minus8.1 \\
& ByT5 & \textbf{\minus1.5} & \textbf{\minus0.5} & \textbf{\minus1.7} \\

\midrule

\multirow{2}{*} {Random Case}
& mT5 & \minus25.7 & \minus14.3 & \minus19.2 \\
& ByT5 & \textbf{\minus1.5} & \textbf{\minus0.2} & \textbf{\minus5.9}\\

\bottomrule

\end{tabular}}
\caption{Degradation of mT5 and ByT5 under various types of noise. ``Clean'' shows original task performance. Subsequent rows show the delta from ``clean'' when adding different types of noise. Learnable noise is added in training and eval, while unseen noise only affects eval.}
\label{tab:noisy_tasks}
\end{table}

\Cref{tab:noisy_tasks} shows the differing ability of ByT5 and mT5 to adapt to learnable noise. We measure the degradation of the task metric between the clean and noisy settings. We observe that mT5 degrades more in the presence of noise than ByT5, across all noise conditions. In the most extreme contrast, rANdOm CaSE (often used as an affective device on social media\footnote{For example, see \url{https://knowyourmeme.com/memes/mocking-spongebob}.}) is hugely detrimental to mT5, with losses of $\minus25.7$ and $\minus14.3$ points, while ByT5 only drops by $\minus1.5$ and $\minus0.2$ points. ByT5 is also quite robust to UPPERCASE and repetitions.

We also test robustness to noise that is \emph{unseen} during training but injected during evaluation. This is relevant in making models more future-proof as well as more resilient to accidental or adversarial spelling mistakes \cite{pruthi-etal-2019-combating, sun-2020-adv-bert}. We evaluate only XNLI and skip TyDiQA-GoldP in this setting, as it is unreasonable to expect a generative model that was fine-tuned to always copy spans from the context to spontaneously ``undo'' corruptions and predict novel spans. The rightmost column of \cref{tab:noisy_tasks} shows that in this more challenging setting, ByT5 is once again more resilient to noise. While some types of unseen noise like A~N~T~S~P~E~A~K are highly detrimental, ByT5 sees only minor degradations for casing noise.

Our findings echo the results of \citet{durrani-etal-2019-one}, who find that character-level models are more robust to real and synthetic noise than BPE or word-based models, across a range of morphological, syntactic and semantic tagging tasks. The more general conclusion that emerges is that token-free models are more robust to noise across many tasks.

\section{Ablation Study}
\label{sec:ablations}

\begin{table}[t!]
\centering
\resizebox{0.95\columnwidth}{!}{
\footnotesize
\setlength\tabcolsep{3pt}
\begin{tabular}{lcl}
\toprule
Model & Params & Description \\
\midrule
ByT5-Large           & 1.23B           & Baseline ByT5 model \\
mT5-Large            & 1.23B           & Baseline mT5 model \\
\midrule
(a) ByT5-36/12-668M      & 668M            & encoder:36, decoder:12 \\
(b) ByT5-24/24-718M      & 718M            & encoder:24, decoder:24 \\
(c) ByT5-12/36-768M      & 768M            & encoder:12, decoder:36 \\
\midrule
(d) mT5-36/12-1.18B      & 1.18B           & encoder:36, decoder:12 \\
\midrule
(e) ByT5-Large-Span3    & 1.23B           & Mean noise span 3.0 \\
(f) ByT5-Large-Span40   & 1.23B           & Mean noise span 40.0 \\
\midrule
(g) CharT5-36/12-1.23B   & 1.23B           & 47K character vocab \\

\bottomrule
\end{tabular}}
\caption{Models used in our ablation study.}
\label{tab:ablation_models}
\end{table}

To better understand the importance of various design choices, we train ablation models and compare these against our baselines on three tasks: XNLI zero-shot, TyDiQA-GoldP and \mbox{GEM-XSum}. Our baselines and ablations are listed in \cref{tab:ablation_models}. The baselines are the parameter-matched ByT5-Large and mT5-Large models discussed above.

\subsection{Matched Transformer Layer Size}

Model (a)~ByT5-36/12-668M is identical to ByT5-Large except that d\textsubscript{model} and d\textsubscript{ff} are matched to \mbox{mT5-Large}, giving a model with $668$ million parameters, \bettertilde{}$54$\% the size of \mbox{ByT5-Large} and \mbox{mT5-Large}. As seen in \cref{tab:ablation_results}, this model is still competitive, and outperforms the roughly similarly sized \mbox{mT5-Base} by a large margin (cf.~\cref{tab:xtreme}). This is evidence that the value of ByT5 does not come solely from using wider transformer layers.

\begin{table}[t!]
\centering
\resizebox{\columnwidth}{!}{
\footnotesize
\setlength\tabcolsep{3pt}
\begin{tabular}{lccc}
\toprule
& XNLI & TyDiQA- & GEM-XSum \\
Model & (Accuracy) & GoldP (F1) & (BLEU) \\
\midrule
ByT5-Large (1.23B) & 79.7 & 87.7 & 11.5 \\
mT5-Large (1.23B) & 81.1 & 85.3 & 10.1 \\
\midrule
(a) ByT5-36/12-668M & 78.3 & 87.8 & 12.3 \\
(b) ByT5-24/24-718M & 75.4 & 83.0 & 7.1 \\
(c) ByT5-12/36-768M & 73.5 & 83.1 & 8.3 \\
\midrule
(d) mT5-36/12-1.18B & 81.5 & 87.1 & 10.8 \\
\midrule
(e) ByT5-Large-Span3 & 79.4 & 87.4 & 10.2 \\
(f) ByT5-Large-Span40 & 78.9 & 88.3 & 12.6 \\
\midrule
(g) CharT5-36/12-1.23B & 79.0 & 87.6 & 11.2 \\
\bottomrule
\end{tabular}}
\caption{Ablation model results across three tasks.}
\label{tab:ablation_results}
\end{table}

\subsection{Encoder/Decoder Balance}

To investigate the effect of decoupling encoder and decoder depth, we train two additional ByT5 models with d\textsubscript{model} and d\textsubscript{ff} matched to \mbox{mT5-Large}: (b)~ByT5-24/24-718M, a ``balanced'' model with 24/24 encoder/decoder layers, and (c)~ByT5-12/36-768M, a ``heavy decoder'' model. As decoder layers have extra parameters used for decoder-encoder attention, these models are bigger than our default heavy encoder setup. Yet despite the extra parameters, these configurations underperform on all tasks, including even the generative \mbox{GEM-XSum} task that we might expect to benefit from a stronger decoder.

To test whether a heavier encoder benefits mT5 as well, we train (d)~mT5-36/12-1.18B, a model with the same configuration as mT5-Large, but switching to 36/12 encoder/decoder layers. As with ByT5, we observe benefits across all three tasks. However, the gains ($+0.4$, $+1.8$, $+0.7$) are much smaller than those of ByT5 ($+2.9$, $+4.8$, $+5.2$).

We suspect a heavy encoder may be particularly important in vocabulary-free models as the encoder stack must stand in for the missing high-capacity token embedding matrix, allowing the model to learn a ``soft lexicon'' covering potentially millions of idiosyncratic mappings from word forms to meanings.
In concurrent work, \citet{wies-2021-transformer} also observe that models with tiny vocabularies benefit from additional depth. One reason the decoder may not need as much capacity is that in inference, the decoder is run autoregressively, using a full forward pass for every token prediction. Given the increased resolution of byte sequences, this means ByT5 predictions will benefit from $2$--$9$ times more passes through the decoder stack depending on the language (see \cref{fig:compression-rates}), as compared to mT5. In this light, even a shallower byte decoder may be sufficient to compete with a larger subword decoder.

\subsection{Masked Span Length}

The T5 \textit{mean span length} hyperparameter controls the average length of the masked spans used in the unsupervised pre-training objective. For T5 and mT5, this was $3$ SentencePiece tokens. For ByT5, we hypothesize that predicting such short byte-spans would be too easy of a task, as this would often just require reconstructing part of a single word (regardless of language). Our final ByT5 models use mean span length of $20$ bytes, which results in more challenging reconstruction tasks. We also show ablations (e--f) with span length $3$ and $40$. \Cref{tab:ablation_results} shows that our baseline with length $20$ performs the best on the classification task XNLI, whereas length $40$ performs better on TyDiQA-GoldP and \mbox{GEM-XSum}, both of which require generating a natural language text output.

\subsection{Character Vocabulary}

A character-level vocabulary serves as an intermediate point between a large subword vocabulary and a tiny byte vocabulary. As a point of comparison, we train (g) CharT5-36/12-1.23B: a model with a vocabulary of $47{,}198$ characters, the same encoder/decoder ratio as ByT5, and the same overall parameter count as ByT5-Large and mT5-Large. To achieve this matched parameter count, we set d\textsubscript{model}=$1376$ and d\textsubscript{ff}=$3840$. The resulting proportion of vocab-related parameters is $11$\% (compared to $42$\% for mT5-Large and $0.06$\% for ByT5-Large). The vocabulary itself is implemented using the SentencePiece library, but with an added restriction that tokens may only represent single characters. The characters cover all those seen in a sample of $4$ million documents taken from the mC4 pre-training corpus, mixing languages with the ratios used during pre-training. We use the \textit{byte-level fallback} mechanism, so no character is out-of-vocabulary.

\Cref{tab:ablation_results} shows that CharT5 is fairly competitive, but performs slightly worse than ByT5 on all three tasks. We suspect this may be due to two factors: (i) CharT5 reserves a capacity for rare characters, and these parameters would be better allocated in the transformer layers, and (ii) using UTF-8 bytes increases the sequence length for non-ASCII text, resulting in extra computational budget for encoding and decoding languages with non-Latin scripts.

\section{Speed Comparisons}
\label{sec:speed}

\begin{table}[t!]
\centering
\resizebox{0.9\columnwidth}{!}{
\footnotesize
\begin{tabular}{lrrrr}
\toprule
      & \multicolumn{2}{c}{sequences / sec} &\multicolumn{2}{c}{einsum ops $\times1e^{12}$} \\
      \cmidrule(lr){2-3} \cmidrule(lr){4-5}
      & mT5 & ByT5\hspace{2.5em} & mT5 & ByT5\hspace{2.5em} \\
\midrule
Small & 1646 & 1232 ($0.75\times$) & 87 & 98 ($1.13\times$) \\
Base & 747 & 576 ($0.77\times$) & 168 & 194 ($1.15\times$) \\
Large & 306 & 232 ($0.76\times$) & 346 & 416 ($1.20\times$) \\
XL & 94 & 70 ($0.74\times$) & 1000 & 1220 ($1.22\times$) \\
XXL & 33 & 25 ($0.76\times$) & 1660 & 2070 ($1.25\times$) \\
\bottomrule 
\end{tabular}}
\caption{Pre-training speed and computation of mT5 vs.~ByT5. \textbf{Left}: Sequences per second pre-training on a TPUv3-64 device. \textbf{Right}: Total einsum operations for a forward pass, as logged by the T5 framework.}
\label{tab:speed_pretraining}

\end{table}

\Cref{tab:speed_pretraining} compares the \textit{pre-training} FLOPs of ByT5 vs.~mT5, as well as the pre-training speed on fixed hardware, as sequences per second with sequence length of $1024$. Across all model sizes, ByT5 requires \bettertilde{}$1.2\times$ more operations, resulting in \bettertilde{}$0.75\times$ as many sequences per second.

\begin{table}[t!]
\centering
\resizebox{0.9\columnwidth}{!}{
\footnotesize
\begin{tabular}{lrrrr}
\toprule
     & \multicolumn{2}{c}{Grapheme-to-Phoneme} &\multicolumn{2}{c}{Dakshina} \\
      \cmidrule(lr){2-3} \cmidrule(lr){4-5}
      & mT5 & ByT5\hspace{1.5em} & mT5 & ByT5\hspace{1.5em} \\
\midrule
Small & 1223 & 1190 (1.0$\times$) & 9483 & 6482 (1.5$\times$) \\
Base  & 726  & 932 (0.8$\times$)  & 7270 & 4272 (1.7$\times$) \\
Large & 387  & 478 (0.8$\times$)  & 4243 & 2282 (1.9$\times$) \\
XL    & 280  & 310 (0.9$\times$)  & 2922 & 1263 (2.3$\times$) \\
XXL   & 150  & 146 (1.0$\times$)  & 1482 & 581 (2.6$\times$)  \\
\midrule
\midrule
     & \multicolumn{2}{c}{XNLI} &\multicolumn{2}{c}{GEM-XSum} \\
      \cmidrule(lr){2-3} \cmidrule(lr){4-5}
      & mT5 & ByT5\hspace{1.5em} & mT5 & ByT5\hspace{1.5em} \\
\midrule
Small & 8632 & 1339 (6.4$\times$) & 750 & 202 (3.7$\times$) \\
Base  & 5157 & 687 (7.5$\times$)  & 450 & 114 (3.9$\times$) \\
Large & 1598 & 168 (9.5$\times$)  & 315 & 51 (6.2$\times$)  \\
XL    & 730  & 81 (9.0$\times$)   & 162 & 25 (6.4$\times$)  \\
XXL   & 261  & 33 (8.0$\times$)   & 61  & 10 (6.3$\times$) \\ 
\bottomrule 

\end{tabular}
}
\caption{Average inference examples per second on the test sets of word-level tasks (top) and sentence- or document-level tasks (bottom). We use a TPUv3-128 for GEM-XSum, and a TPUv3-32 elsewhere.}
\label{tab:speed_inference}

\end{table}

\Cref{tab:speed_inference} compares the \textit{inference} speed of ByT5 and mT5 by measuring the average number of inference predictions per second across four tasks. On word-level tasks, ByT5 is fairly competitive: on \textsc{Sigmorphon} 2020 Grapheme-to-Phoneme, where targets are written using the International Phonetic Alphabet, ByT5 and mT5 have similar inference speed; on Dakshina transliteration, ByT5 is $1.5$ to $2.6$ times slower. On tasks with longer input sequences, the slowdown is more pronounced: on \mbox{GEM-XSum}\footnote{To stay within reasonable memory requirements for the XXL models, we filter out \mbox{GEM-XSum} examples with inputs longer than 8192 characters (less than 1\% of the data).} (document summarization), ByT5 is $3.7$ to $6.4$ times slower than mT5, while on XNLI zero-shot classification it is $6.4$ to $9.5$ times slower. More generally, we observe that---as expected due to its deeper encoder and shallower decoder---ByT5 achieves more competitive inference speed (relative to mT5) on tasks with short inputs and/or long targets. In this light, XNLI represents something of a worst-case, where inputs are sentence pairs and labels are single digits \{0,~1,~2\}.

The time required for \textit{fine-tuning} is also variable across tasks. When holding batch size constant at a fixed number of tokens, we find that ByT5 typically takes more fine-tuning steps than mT5 to reach optimal performance on a holdout set. For example, ByT5-Large took $1.2\times$ as many steps as mT5-Large to reach peak validation performance on XNLI zero-shot, $2.6\times$ as many steps for TyDiQA-GoldP, and $4.5\times$ as many for \mbox{GEM-XSum}. This overall trend is expected, in that fewer labeled examples fit into each ByT5 fine-tuning batch. However, on tasks that strongly favor byte-level representations, ByT5 reaches peak performance in \emph{fewer} fine-tuning steps, suggesting that the model can generalize better from a small number of training examples. For example, ByT5-Large took $2.5\times$ fewer steps than mT5-Large to reach peak performance on Dakshina.

Overall, we believe that the additional pre-training cost (roughly $+33$\% wall time) and the additional fine-tuning cost (for some tasks) is justified in non-latency-sensitive applications by the benefits of reduced system complexity, better robustness to noise, and improved task performance on many benchmarks.

\section{Conclusion}
\label{sec:conclusion}

In this work, we presented ByT5, a token-free variant of multilingual T5 \cite{mt5} that simplifies the NLP pipeline by doing away with vocabulary building, text preprocessing and tokenization. On downstream task quality, ByT5 is competitive with parameter-matched mT5 models that rely on SentencePiece vocabulary. Specifically, ByT5 outperforms mT5 in any of these five scenarios: (1)~at model sizes under $1$ billion parameters, (2)~on generative tasks, (3)~on multilingual tasks with in-language labels, (4)~on word-level tasks sensitive to spelling and/or pronunciation, and (5)~in the presence of various types of noise.

While beating mT5 in many cases, ByT5 slightly underperformed in certain conditions---most notably, on English classification tasks for model sizes over $1$ billion parameters. In future work, it will also be important to evaluate token-free approaches on a more diverse set of tasks, especially those where character-based models have traditionally struggled. These include word similarity tasks \cite{hiebert-etal-2018-interpreting}, syntactic and semantic tagging tasks \cite{durrani-etal-2019-one}, and machine translation from a non-English source into English \cite{shaham-levy-2021-neural}.

Through ablations, we showed that byte-level encoder-decoder models benefit from a ``heavier'' encoder (decoupling encoder and decoder depth), and that the pre-training task benefits from masking longer ID sequences. We also showed that for fixed parameter count, character-level models give similar but somewhat worse results.

Interestingly, the gains we observe with ByT5 are achieved \emph{despite} the model being pre-trained on $4\times$ less text than mT5. This suggests that byte-level models may be more data efficient learners.

These gains in design simplicity, task quality and data efficiency come at the cost of additional computation. Our ``hands-off'' approach of feeding raw UTF-8 bytes directly into the Transformer costs $+33\%$ pre-training time, as well as longer inference time (up to $10\times$ slower in the worst case). As such, there is significant room for improvement. We believe techniques such as hash embeddings, local attention and down-sampling \cite{clark2021canine}, as well as sparse computation \cite{fedus-2021-switch} can help address latency issues, removing the remaining barriers to a token-free future.

\subsection*{Acknowledgements}

We thank Jon Clark and Dan Garrette for discussion around token-free approaches and Noam Shazeer for help around model parallelism in T5. We also thank Jon Clark and the TACL reviewers and action editors for helpful comments on an earlier draft.

\bibliography{anthology,tacl2018v2}

\begin{thebibliography}{64}
\expandafter\ifx\csname natexlab\endcsname\relax\def\natexlab#1{#1}\fi

\bibitem[{Akbik et~al.(2018)Akbik, Blythe, and
  Vollgraf}]{akbik-etal-2018-contextual}
Alan Akbik, Duncan Blythe, and Roland Vollgraf. 2018.
\newblock \href {https://www.aclweb.org/anthology/C18-1139} {Contextual string
  embeddings for sequence labeling}.
\newblock In \emph{Proceedings of the 27th International Conference on
  Computational Linguistics}, pages 1638--1649, Santa Fe, New Mexico, USA.
  Association for Computational Linguistics.

\bibitem[{Al-Rfou et~al.(2019)Al-Rfou, Choe, Constant, Guo, and
  Jones}]{al-rfou-2019-character-lm}
Rami Al-Rfou, Dokook Choe, Noah Constant, Mandy Guo, and Llion Jones. 2019.
\newblock \href {https://doi.org/10.1609/aaai.v33i01.33013159} {Character-level
  language modeling with deeper self-attention}.
\newblock \emph{Proceedings of the AAAI Conference on Artificial Intelligence},
  33(01):3159--3166.

\bibitem[{Artetxe et~al.(2020)Artetxe, Ruder, and
  Yogatama}]{artetxe-etal-2020-cross}
Mikel Artetxe, Sebastian Ruder, and Dani Yogatama. 2020.
\newblock \href {https://doi.org/10.18653/v1/2020.acl-main.421} {On the
  cross-lingual transferability of monolingual representations}.
\newblock In \emph{Proceedings of the 58th Annual Meeting of the Association
  for Computational Linguistics}, pages 4623--4637, Online. Association for
  Computational Linguistics.

\bibitem[{Caswell et~al.(2020)Caswell, Breiner, van Esch, and
  Bapna}]{caswell2020language}
Isaac Caswell, Theresa Breiner, Daan van Esch, and Ankur Bapna. 2020.
\newblock \href {https://doi.org/10.18653/v1/2020.coling-main.579} {Language
  {ID} in the wild: Unexpected challenges on the path to a thousand-language
  web text corpus}.
\newblock In \emph{Proceedings of the 28th International Conference on
  Computational Linguistics}, pages 6588--6608, Barcelona, Spain (Online).
  International Committee on Computational Linguistics.

\bibitem[{Chen et~al.(2020)Chen, Xu, Cheng, Xiaochuan, Zhang, Song, Wang, Qi,
  and Chu}]{chen2020question}
Kunlong Chen, Weidi Xu, Xingyi Cheng, Zou Xiaochuan, Yuyu Zhang, Le~Song,
  Taifeng Wang, Yuan Qi, and Wei Chu. 2020.
\newblock \href {http://arxiv.org/abs/2009.07448} {Question directed graph
  attention network for numerical reasoning over text}.

\bibitem[{Cherry et~al.(2018)Cherry, Foster, Bapna, Firat, and
  Macherey}]{cherry-etal-2018-revisiting}
Colin Cherry, George Foster, Ankur Bapna, Orhan Firat, and Wolfgang Macherey.
  2018.
\newblock \href {https://doi.org/10.18653/v1/D18-1461} {Revisiting
  character-based neural machine translation with capacity and compression}.
\newblock In \emph{Proceedings of the 2018 Conference on Empirical Methods in
  Natural Language Processing}, pages 4295--4305, Brussels, Belgium.
  Association for Computational Linguistics.

\bibitem[{Choe et~al.(2019)Choe, Al{-}Rfou, Guo, Lee, and
  Constant}]{Choe2019BridgingTG}
Dokook Choe, Rami Al{-}Rfou, Mandy Guo, Heeyoung Lee, and Noah Constant. 2019.
\newblock \href {http://arxiv.org/abs/1908.10322} {Bridging the gap for
  tokenizer-free language models}.
\newblock \emph{CoRR}, abs/1908.10322v1.

\bibitem[{Chung et~al.(2017)Chung, Ahn, and Bengio}]{chung-2017-hierarchical}
Junyoung Chung, Sungjin Ahn, and Yoshua Bengio. 2017.
\newblock \href {https://openreview.net/forum?id=S1di0sfgl} {Hierarchical
  multiscale recurrent neural networks}.
\newblock In \emph{5th International Conference on Learning Representations,
  {ICLR} 2017, Toulon, France, April 24-26, 2017, Conference Track
  Proceedings}. OpenReview.net.

\bibitem[{Chung et~al.(2016)Chung, Cho, and Bengio}]{chung-etal-2016-character}
Junyoung Chung, Kyunghyun Cho, and Yoshua Bengio. 2016.
\newblock \href {https://doi.org/10.18653/v1/P16-1160} {A character-level
  decoder without explicit segmentation for neural machine translation}.
\newblock In \emph{Proceedings of the 54th Annual Meeting of the Association
  for Computational Linguistics (Volume 1: Long Papers)}, pages 1693--1703,
  Berlin, Germany. Association for Computational Linguistics.

\bibitem[{Clark et~al.(2020)Clark, Choi, Collins, Garrette, Kwiatkowski,
  Nikolaev, and Palomaki}]{clark-etal-2020-tydi}
Jonathan~H. Clark, Eunsol Choi, Michael Collins, Dan Garrette, Tom Kwiatkowski,
  Vitaly Nikolaev, and Jennimaria Palomaki. 2020.
\newblock \href {https://doi.org/10.1162/tacl_a_00317} {{T}y{D}i {QA}: A
  benchmark for information-seeking question answering in typologically diverse
  languages}.
\newblock \emph{Transactions of the Association for Computational Linguistics},
  8:454--470.

\bibitem[{Clark et~al.(2021)Clark, Garrette, Turc, and
  Wieting}]{clark2021canine}
Jonathan~H. Clark, Dan Garrette, Iulia Turc, and John Wieting. 2021.
\newblock \href {http://arxiv.org/abs/2103.06874} {{CANINE:} pre-training an
  efficient tokenization-free encoder for language representation}.
\newblock \emph{CoRR}, abs/2103.06874v3.

\bibitem[{Conneau et~al.(2018)Conneau, Rinott, Lample, Williams, Bowman,
  Schwenk, and Stoyanov}]{conneau2018xnli}
Alexis Conneau, Ruty Rinott, Guillaume Lample, Adina Williams, Samuel Bowman,
  Holger Schwenk, and Veselin Stoyanov. 2018.
\newblock \href {https://doi.org/10.18653/v1/D18-1269} {{XNLI}: Evaluating
  cross-lingual sentence representations}.
\newblock In \emph{Proceedings of the 2018 Conference on Empirical Methods in
  Natural Language Processing}, pages 2475--2485, Brussels, Belgium.
  Association for Computational Linguistics.

\bibitem[{Costa-juss{\`a} et~al.(2017)Costa-juss{\`a}, Escolano, and
  Fonollosa}]{costa-jussa-etal-2017-byte}
Marta~R. Costa-juss{\`a}, Carlos Escolano, and Jos{\'e} A.~R. Fonollosa. 2017.
\newblock \href {https://doi.org/10.18653/v1/W17-4123} {Byte-based neural
  machine translation}.
\newblock In \emph{Proceedings of the First Workshop on Subword and Character
  Level Models in {NLP}}, pages 154--158, Copenhagen, Denmark. Association for
  Computational Linguistics.

\bibitem[{Dua et~al.(2019)Dua, Wang, Dasigi, Stanovsky, Singh, and
  Gardner}]{dua-etal-2019-drop}
Dheeru Dua, Yizhong Wang, Pradeep Dasigi, Gabriel Stanovsky, Sameer Singh, and
  Matt Gardner. 2019.
\newblock \href {https://doi.org/10.18653/v1/N19-1246} {{DROP}: A reading
  comprehension benchmark requiring discrete reasoning over paragraphs}.
\newblock In \emph{Proceedings of the 2019 Conference of the North {A}merican
  Chapter of the Association for Computational Linguistics: Human Language
  Technologies, Volume 1 (Long and Short Papers)}, pages 2368--2378,
  Minneapolis, Minnesota. Association for Computational Linguistics.

\bibitem[{Durrani et~al.(2019)Durrani, Dalvi, Sajjad, Belinkov, and
  Nakov}]{durrani-etal-2019-one}
Nadir Durrani, Fahim Dalvi, Hassan Sajjad, Yonatan Belinkov, and Preslav Nakov.
  2019.
\newblock \href {https://doi.org/10.18653/v1/N19-1154} {One size does not fit
  all: Comparing {NMT} representations of different granularities}.
\newblock In \emph{Proceedings of the 2019 Conference of the North {A}merican
  Chapter of the Association for Computational Linguistics: Human Language
  Technologies, Volume 1 (Long and Short Papers)}, pages 1504--1516,
  Minneapolis, Minnesota. Association for Computational Linguistics.

\bibitem[{El~Boukkouri et~al.(2020)El~Boukkouri, Ferret, Lavergne, Noji,
  Zweigenbaum, and Tsujii}]{el-boukkouri-etal-2020-characterbert}
Hicham El~Boukkouri, Olivier Ferret, Thomas Lavergne, Hiroshi Noji, Pierre
  Zweigenbaum, and Jun{'}ichi Tsujii. 2020.
\newblock \href {https://doi.org/10.18653/v1/2020.coling-main.609}
  {{C}haracter{BERT}: Reconciling {ELM}o and {BERT} for word-level
  open-vocabulary representations from characters}.
\newblock In \emph{Proceedings of the 28th International Conference on
  Computational Linguistics}, pages 6903--6915, Barcelona, Spain (Online).
  International Committee on Computational Linguistics.

\bibitem[{Fedus et~al.(2021)Fedus, Zoph, and Shazeer}]{fedus-2021-switch}
William Fedus, Barret Zoph, and Noam Shazeer. 2021.
\newblock \href {http://arxiv.org/abs/2101.03961} {Switch transformers: Scaling
  to trillion parameter models with simple and efficient sparsity}.
\newblock \emph{CoRR}, abs/2101.03961v1.

\bibitem[{Garcia et~al.(2021)Garcia, Constant, Parikh, and
  Firat}]{garcia-2021-continual}
Xavier Garcia, Noah Constant, Ankur Parikh, and Orhan Firat. 2021.
\newblock \href {https://doi.org/10.18653/v1/2021.naacl-main.93} {Towards
  continual learning for multilingual machine translation via vocabulary
  substitution}.
\newblock In \emph{Proceedings of the 2021 Conference of the North American
  Chapter of the Association for Computational Linguistics: Human Language
  Technologies}, pages 1184--1192, Online. Association for Computational
  Linguistics.

\bibitem[{Gehrmann et~al.(2021)Gehrmann, Adewumi, Aggarwal, Ammanamanchi,
  Anuoluwapo, Bosselut, Chandu, Clinciu, Das, Dhole, Du, Durmus, Dusek, Emezue,
  Gangal, Garbacea, Hashimoto, Hou, Jernite, Jhamtani, Ji, Jolly, Kumar,
  Ladhak, Madaan, Maddela, Mahajan, Mahamood, Majumder, Martins,
  McMillan{-}Major, Mille, van Miltenburg, Nadeem, Narayan, Nikolaev,
  Niyongabo, Osei, Parikh, Perez{-}Beltrachini, Rao, Raunak, Rodriguez,
  Santhanam, Sedoc, Sellam, Shaikh, Shimorina, Cabezudo, Strobelt, Subramani,
  Xu, Yang, Yerukola, and Zhou}]{gehrmann2021gem}
Sebastian Gehrmann, Tosin~P. Adewumi, Karmanya Aggarwal, Pawan~Sasanka
  Ammanamanchi, Aremu Anuoluwapo, Antoine Bosselut, Khyathi~Raghavi Chandu,
  Miruna{-}Adriana Clinciu, Dipanjan Das, Kaustubh~D. Dhole, Wanyu Du, Esin
  Durmus, Ondrej Dusek, Chris Emezue, Varun Gangal, Cristina Garbacea,
  Tatsunori Hashimoto, Yufang Hou, Yacine Jernite, Harsh Jhamtani, Yangfeng Ji,
  Shailza Jolly, Dhruv Kumar, Faisal Ladhak, Aman Madaan, Mounica Maddela,
  Khyati Mahajan, Saad Mahamood, Bodhisattwa~Prasad Majumder, Pedro~Henrique
  Martins, Angelina McMillan{-}Major, Simon Mille, Emiel van Miltenburg, Moin
  Nadeem, Shashi Narayan, Vitaly Nikolaev, Rubungo~Andre Niyongabo, Salomey
  Osei, Ankur~P. Parikh, Laura Perez{-}Beltrachini, Niranjan~Ramesh Rao, Vikas
  Raunak, Juan~Diego Rodriguez, Sashank Santhanam, Jo{\~{a}}o Sedoc, Thibault
  Sellam, Samira Shaikh, Anastasia Shimorina, Marco Antonio~Sobrevilla
  Cabezudo, Hendrik Strobelt, Nishant Subramani, Wei Xu, Diyi Yang, Akhila
  Yerukola, and Jiawei Zhou. 2021.
\newblock \href {http://arxiv.org/abs/2102.01672} {The {GEM} benchmark: Natural
  language generation, its evaluation and metrics}.
\newblock \emph{CoRR}, abs/2102.01672v3.

\bibitem[{Gillick et~al.(2016)Gillick, Brunk, Vinyals, and
  Subramanya}]{gillick-etal-2016-multilingual}
Dan Gillick, Cliff Brunk, Oriol Vinyals, and Amarnag Subramanya. 2016.
\newblock \href {https://doi.org/10.18653/v1/N16-1155} {Multilingual language
  processing from bytes}.
\newblock In \emph{Proceedings of the 2016 Conference of the North {A}merican
  Chapter of the Association for Computational Linguistics: Human Language
  Technologies}, pages 1296--1306, San Diego, California. Association for
  Computational Linguistics.

\bibitem[{Gorman et~al.(2020)Gorman, Ashby, Goyzueta, McCarthy, Wu, and
  You}]{gorman-etal-2020-sigmorphon}
Kyle Gorman, Lucas~F.E. Ashby, Aaron Goyzueta, Arya McCarthy, Shijie Wu, and
  Daniel You. 2020.
\newblock \href {https://doi.org/10.18653/v1/2020.sigmorphon-1.2} {The
  {SIGMORPHON} 2020 shared task on multilingual grapheme-to-phoneme
  conversion}.
\newblock In \emph{Proceedings of the 17th SIGMORPHON Workshop on Computational
  Research in Phonetics, Phonology, and Morphology}, pages 40--50, Online.
  Association for Computational Linguistics.

\bibitem[{Graves(2013)}]{graves-2013-generating}
Alex Graves. 2013.
\newblock \href {http://arxiv.org/abs/1308.0850} {Generating sequences with
  recurrent neural networks}.
\newblock \emph{CoRR}, abs/1308.0850v5.

\bibitem[{Graves(2016)}]{graves-2016-adaptive}
Alex Graves. 2016.
\newblock \href {http://arxiv.org/abs/1603.08983} {Adaptive computation time
  for recurrent neural networks}.
\newblock \emph{CoRR}, abs/1603.08983v6.

\bibitem[{Ha et~al.(2017)Ha, Dai, and Le}]{ha-2017-hypernets}
David Ha, Andrew~M. Dai, and Quoc~V. Le. 2017.
\newblock \href {https://openreview.net/forum?id=rkpACe1lx} {Hypernetworks}.
\newblock In \emph{5th International Conference on Learning Representations,
  {ICLR} 2017, Toulon, France, April 24-26, 2017, Conference Track
  Proceedings}. OpenReview.net.

\bibitem[{Hiebert et~al.(2018)Hiebert, Peterson, Fyshe, and
  Mehta}]{hiebert-etal-2018-interpreting}
Avery Hiebert, Cole Peterson, Alona Fyshe, and Nishant Mehta. 2018.
\newblock \href {https://doi.org/10.18653/v1/W18-5428} {Interpreting word-level
  hidden state behaviour of character-level {LSTM} language models}.
\newblock In \emph{Proceedings of the 2018 {EMNLP} Workshop {B}lackbox{NLP}:
  Analyzing and Interpreting Neural Networks for {NLP}}, pages 258--266,
  Brussels, Belgium. Association for Computational Linguistics.

\bibitem[{Hu et~al.(2020)Hu, Ruder, Siddhant, Neubig, Firat, and
  Johnson}]{hu_2020_xtreme}
Junjie Hu, Sebastian Ruder, Aditya Siddhant, Graham Neubig, Orhan Firat, and
  Melvin Johnson. 2020.
\newblock \href {http://proceedings.mlr.press/v119/hu20b.html} {{XTREME}: A
  massively multilingual multi-task benchmark for evaluating cross-lingual
  generalisation}.
\newblock In \emph{Proceedings of the 37th International Conference on Machine
  Learning}, volume 119 of \emph{Proceedings of Machine Learning Research},
  pages 4411--4421. PMLR.

\bibitem[{J{\'{o}}zefowicz et~al.(2016)J{\'{o}}zefowicz, Vinyals, Schuster,
  Shazeer, and Wu}]{rafal-2016-exploring}
Rafal J{\'{o}}zefowicz, Oriol Vinyals, Mike Schuster, Noam Shazeer, and Yonghui
  Wu. 2016.
\newblock \href {http://arxiv.org/abs/1602.02410} {Exploring the limits of
  language modeling}.
\newblock \emph{CoRR}, abs/1602.02410v2.

\bibitem[{Kalchbrenner et~al.(2016)Kalchbrenner, Espeholt, Simonyan, van~den
  Oord, Graves, and Kavukcuoglu}]{kalchbrenner-2016-bytenet}
Nal Kalchbrenner, Lasse Espeholt, Karen Simonyan, A{\"{a}}ron van~den Oord,
  Alex Graves, and Koray Kavukcuoglu. 2016.
\newblock \href {http://arxiv.org/abs/1610.10099} {Neural machine translation
  in linear time}.
\newblock \emph{CoRR}, abs/1610.10099v2.

\bibitem[{Kaplan et~al.(2020)Kaplan, McCandlish, Henighan, Brown, Chess, Child,
  Gray, Radford, Wu, and Amodei}]{kaplan2020scaling}
Jared Kaplan, Sam McCandlish, Tom Henighan, Tom~B. Brown, Benjamin Chess, Rewon
  Child, Scott Gray, Alec Radford, Jeffrey Wu, and Dario Amodei. 2020.
\newblock \href {http://arxiv.org/abs/2001.08361} {Scaling laws for neural
  language models}.
\newblock \emph{CoRR}, abs/2001.08361v1.

\bibitem[{Kim et~al.(2016)Kim, Jernite, Sontag, and
  Rush}]{kim-2016-character-aware}
Yoon Kim, Yacine Jernite, David Sontag, and Alexander~M. Rush. 2016.
\newblock Character-aware neural language models.
\newblock In \emph{Proceedings of the Thirtieth AAAI Conference on Artificial
  Intelligence}, AAAI'16, page 2741–2749. AAAI Press.

\bibitem[{Kudo(2018)}]{kudo-2018-subword}
Taku Kudo. 2018.
\newblock \href {https://doi.org/10.18653/v1/P18-1007} {Subword regularization:
  Improving neural network translation models with multiple subword
  candidates}.
\newblock In \emph{Proceedings of the 56th Annual Meeting of the Association
  for Computational Linguistics (Volume 1: Long Papers)}, pages 66--75,
  Melbourne, Australia. Association for Computational Linguistics.

\bibitem[{Kudo and Richardson(2018)}]{kudo-richardson-2018-sentencepiece}
Taku Kudo and John Richardson. 2018.
\newblock \href {https://doi.org/10.18653/v1/D18-2012} {{S}entence{P}iece: A
  simple and language independent subword tokenizer and detokenizer for neural
  text processing}.
\newblock In \emph{Proceedings of the 2018 Conference on Empirical Methods in
  Natural Language Processing: System Demonstrations}, pages 66--71, Brussels,
  Belgium. Association for Computational Linguistics.

\bibitem[{Lee et~al.(2017)Lee, Cho, and Hofmann}]{lee-etal-2017-fully}
Jason Lee, Kyunghyun Cho, and Thomas Hofmann. 2017.
\newblock \href {https://doi.org/10.1162/tacl_a_00067} {Fully character-level
  neural machine translation without explicit segmentation}.
\newblock \emph{Transactions of the Association for Computational Linguistics},
  5:365--378.

\bibitem[{Lewis et~al.(2020)Lewis, Oguz, Rinott, Riedel, and
  Schwenk}]{lewis-etal-2020-mlqa}
Patrick Lewis, Barlas Oguz, Ruty Rinott, Sebastian Riedel, and Holger Schwenk.
  2020.
\newblock \href {https://doi.org/10.18653/v1/2020.acl-main.653} {{MLQA}:
  Evaluating cross-lingual extractive question answering}.
\newblock In \emph{Proceedings of the 58th Annual Meeting of the Association
  for Computational Linguistics}, pages 7315--7330, Online. Association for
  Computational Linguistics.

\bibitem[{Li et~al.(2019)Li, Zhang, Sainath, Wu, and Chan}]{li-2019-bytes}
Bo~Li, Yu~Zhang, Tara Sainath, Yonghui Wu, and William Chan. 2019.
\newblock \href {https://doi.org/10.1109/ICASSP.2019.8682674} {Bytes are all
  you need: End-to-end multilingual speech recognition and synthesis with
  bytes}.
\newblock In \emph{ICASSP 2019 - 2019 IEEE International Conference on
  Acoustics, Speech and Signal Processing (ICASSP)}, pages 5621--5625.

\bibitem[{Ling et~al.(2015)Ling, Trancoso, Dyer, and
  Black}]{ling-2015-character}
Wang Ling, Isabel Trancoso, Chris Dyer, and Alan~W. Black. 2015.
\newblock \href {http://arxiv.org/abs/1511.04586} {Character-based neural
  machine translation}.
\newblock \emph{CoRR}, abs/1511.04586v1.

\bibitem[{Melis et~al.(2018)Melis, Dyer, and Blunsom}]{melis-2018-sota-eval}
G{\'{a}}bor Melis, Chris Dyer, and Phil Blunsom. 2018.
\newblock \href {https://openreview.net/forum?id=ByJHuTgA-} {On the state of
  the art of evaluation in neural language models}.
\newblock In \emph{6th International Conference on Learning Representations,
  {ICLR} 2018, Vancouver, BC, Canada, April 30 - May 3, 2018, Conference Track
  Proceedings}. OpenReview.net.

\bibitem[{Narayan et~al.(2018)Narayan, Cohen, and
  Lapata}]{narayan-etal-2018-dont}
Shashi Narayan, Shay~B. Cohen, and Mirella Lapata. 2018.
\newblock \href {https://doi.org/10.18653/v1/D18-1206} {Don{'}t give me the
  details, just the summary! topic-aware convolutional neural networks for
  extreme summarization}.
\newblock In \emph{Proceedings of the 2018 Conference on Empirical Methods in
  Natural Language Processing}, pages 1797--1807, Brussels, Belgium.
  Association for Computational Linguistics.

\bibitem[{Pan et~al.(2017)Pan, Zhang, May, Nothman, Knight, and
  Ji}]{pan-etal-2017-cross}
Xiaoman Pan, Boliang Zhang, Jonathan May, Joel Nothman, Kevin Knight, and Heng
  Ji. 2017.
\newblock \href {https://doi.org/10.18653/v1/P17-1178} {Cross-lingual name
  tagging and linking for 282 languages}.
\newblock In \emph{Proceedings of the 55th Annual Meeting of the Association
  for Computational Linguistics (Volume 1: Long Papers)}, pages 1946--1958,
  Vancouver, Canada. Association for Computational Linguistics.

\bibitem[{Peters and Martins(2020)}]{peters-martins-2020-one}
Ben Peters and Andr{\'e} F.~T. Martins. 2020.
\newblock \href {https://doi.org/10.18653/v1/2020.sigmorphon-1.4}
  {One-size-fits-all multilingual models}.
\newblock In \emph{Proceedings of the 17th SIGMORPHON Workshop on Computational
  Research in Phonetics, Phonology, and Morphology}, pages 63--69, Online.
  Association for Computational Linguistics.

\bibitem[{Peters et~al.(2018)Peters, Neumann, Iyyer, Gardner, Clark, Lee, and
  Zettlemoyer}]{peters-etal-2018-deep}
Matthew Peters, Mark Neumann, Mohit Iyyer, Matt Gardner, Christopher Clark,
  Kenton Lee, and Luke Zettlemoyer. 2018.
\newblock \href {https://doi.org/10.18653/v1/N18-1202} {Deep contextualized
  word representations}.
\newblock In \emph{Proceedings of the 2018 Conference of the North {A}merican
  Chapter of the Association for Computational Linguistics: Human Language
  Technologies, Volume 1 (Long Papers)}, pages 2227--2237, New Orleans,
  Louisiana. Association for Computational Linguistics.

\bibitem[{Pruthi et~al.(2019)Pruthi, Dhingra, and
  Lipton}]{pruthi-etal-2019-combating}
Danish Pruthi, Bhuwan Dhingra, and Zachary~C. Lipton. 2019.
\newblock \href {https://doi.org/10.18653/v1/P19-1561} {Combating adversarial
  misspellings with robust word recognition}.
\newblock In \emph{Proceedings of the 57th Annual Meeting of the Association
  for Computational Linguistics}, pages 5582--5591, Florence, Italy.
  Association for Computational Linguistics.

\bibitem[{Raffel et~al.(2020)Raffel, Shazeer, Roberts, Lee, Narang, Matena,
  Zhou, Li, and Liu}]{raffel-2020-t5}
Colin Raffel, Noam Shazeer, Adam Roberts, Katherine Lee, Sharan Narang, Michael
  Matena, Yanqi Zhou, Wei Li, and Peter~J. Liu. 2020.
\newblock \href {http://jmlr.org/papers/v21/20-074.html} {Exploring the limits
  of transfer learning with a unified text-to-text transformer}.
\newblock \emph{Journal of Machine Learning Research}, 21(140):1--67.

\bibitem[{Roark et~al.(2020)Roark, Wolf-Sonkin, Kirov, Mielke, Johny,
  Demirsahin, and Hall}]{roark-etal-2020-processing}
Brian Roark, Lawrence Wolf-Sonkin, Christo Kirov, Sabrina~J. Mielke, Cibu
  Johny, Isin Demirsahin, and Keith Hall. 2020.
\newblock \href {https://www.aclweb.org/anthology/2020.lrec-1.294} {Processing
  {S}outh {A}sian languages written in the {L}atin script: the dakshina
  dataset}.
\newblock In \emph{Proceedings of the 12th Language Resources and Evaluation
  Conference}, pages 2413--2423, Marseille, France. European Language Resources
  Association.

\bibitem[{Sennrich et~al.(2016)Sennrich, Haddow, and
  Birch}]{sennrich-etal-2016-neural}
Rico Sennrich, Barry Haddow, and Alexandra Birch. 2016.
\newblock \href {https://doi.org/10.18653/v1/P16-1162} {Neural machine
  translation of rare words with subword units}.
\newblock In \emph{Proceedings of the 54th Annual Meeting of the Association
  for Computational Linguistics (Volume 1: Long Papers)}, pages 1715--1725,
  Berlin, Germany. Association for Computational Linguistics.

\bibitem[{Shaham and Levy(2021)}]{shaham-levy-2021-neural}
Uri Shaham and Omer Levy. 2021.
\newblock \href {https://doi.org/10.18653/v1/2021.naacl-main.17} {Neural
  machine translation without embeddings}.
\newblock In \emph{Proceedings of the 2021 Conference of the North American
  Chapter of the Association for Computational Linguistics: Human Language
  Technologies}, pages 181--186, Online. Association for Computational
  Linguistics.

\bibitem[{Sun et~al.(2020)Sun, Hashimoto, Yin, Asai, Li, Yu, and
  Xiong}]{sun-2020-adv-bert}
Lichao Sun, Kazuma Hashimoto, Wenpeng Yin, Akari Asai, Jia Li, Philip~S. Yu,
  and Caiming Xiong. 2020.
\newblock \href {http://arxiv.org/abs/2003.04985} {Adv-{BERT}: {BERT} is not
  robust on misspellings! {G}enerating nature adversarial samples on {BERT}}.
\newblock \emph{CoRR}, abs/2003.04985v1.

\bibitem[{Sutskever et~al.(2011)Sutskever, Martens, and
  Hinton}]{sutskever-2011-generating}
Ilya Sutskever, James Martens, and Geoffrey Hinton. 2011.
\newblock Generating text with recurrent neural networks.
\newblock In \emph{Proceedings of the 28th International Conference on
  International Conference on Machine Learning}, ICML'11, page 1017–1024,
  Madison, WI, USA. Omnipress.

\bibitem[{Tay et~al.(2020)Tay, Dehghani, Bahri, and Metzler}]{tay2020efficient}
Yi~Tay, Mostafa Dehghani, Dara Bahri, and Donald Metzler. 2020.
\newblock \href {http://arxiv.org/abs/2009.06732} {Efficient transformers: {A}
  survey}.
\newblock \emph{CoRR}, abs/2009.06732v2.

\bibitem[{Vaswani et~al.(2017)Vaswani, Shazeer, Parmar, Uszkoreit, Jones,
  Gomez, Kaiser, and Polosukhin}]{vaswani2017attention}
Ashish Vaswani, Noam Shazeer, Niki Parmar, Jakob Uszkoreit, Llion Jones,
  Aidan~N Gomez, \L{}ukasz Kaiser, and Illia Polosukhin. 2017.
\newblock \href
  {https://proceedings.neurips.cc/paper/2017/file/3f5ee243547dee91fbd053c1c4a845aa-Paper.pdf}
  {Attention is all you need}.
\newblock In \emph{Advances in Neural Information Processing Systems},
  volume~30. Curran Associates, Inc.

\bibitem[{Vylomova et~al.(2020)Vylomova, White, Salesky, Mielke, Wu, Ponti,
  Hall~Maudslay, Zmigrod, Valvoda, Toldova, Tyers, Klyachko, Yegorov,
  Krizhanovsky, Czarnowska, Nikkarinen, Krizhanovsky, Pimentel,
  Torroba~Hennigen, Kirov, Nicolai, Williams, Anastasopoulos, Cruz, Chodroff,
  Cotterell, Silfverberg, and Hulden}]{vylomova-etal-2020-sigmorphon}
Ekaterina Vylomova, Jennifer White, Elizabeth Salesky, Sabrina~J. Mielke,
  Shijie Wu, Edoardo~Maria Ponti, Rowan Hall~Maudslay, Ran Zmigrod, Josef
  Valvoda, Svetlana Toldova, Francis Tyers, Elena Klyachko, Ilya Yegorov,
  Natalia Krizhanovsky, Paula Czarnowska, Irene Nikkarinen, Andrew
  Krizhanovsky, Tiago Pimentel, Lucas Torroba~Hennigen, Christo Kirov, Garrett
  Nicolai, Adina Williams, Antonios Anastasopoulos, Hilaria Cruz, Eleanor
  Chodroff, Ryan Cotterell, Miikka Silfverberg, and Mans Hulden. 2020.
\newblock \href {https://doi.org/10.18653/v1/2020.sigmorphon-1.1} {{SIGMORPHON}
  2020 shared task 0: Typologically diverse morphological inflection}.
\newblock In \emph{Proceedings of the 17th SIGMORPHON Workshop on Computational
  Research in Phonetics, Phonology, and Morphology}, pages 1--39, Online.
  Association for Computational Linguistics.

\bibitem[{Wang et~al.(2019{\natexlab{a}})Wang, Pruksachatkun, Nangia, Singh,
  Michael, Hill, Levy, and Bowman}]{wang2019superglue}
Alex Wang, Yada Pruksachatkun, Nikita Nangia, Amanpreet Singh, Julian Michael,
  Felix Hill, Omer Levy, and Samuel Bowman. 2019{\natexlab{a}}.
\newblock \href
  {https://proceedings.neurips.cc/paper/2019/file/4496bf24afe7fab6f046bf4923da8de6-Paper.pdf}
  {{SuperGLUE}: A stickier benchmark for general-purpose language understanding
  systems}.
\newblock In \emph{Advances in Neural Information Processing Systems},
  volume~32. Curran Associates, Inc.

\bibitem[{Wang et~al.(2019{\natexlab{b}})Wang, Singh, Michael, Hill, Levy, and
  Bowman}]{wang2019glue}
Alex Wang, Amanpreet Singh, Julian Michael, Felix Hill, Omer Levy, and
  Samuel~R. Bowman. 2019{\natexlab{b}}.
\newblock \href {https://openreview.net/forum?id=rJ4km2R5t7} {{GLUE}: A
  multi-task benchmark and analysis platform for natural language
  understanding}.
\newblock In \emph{International Conference on Learning Representations}.

\bibitem[{Wang et~al.(2020)Wang, Cho, and Gu}]{wang-2020-nmt-with-byte}
Changhan Wang, Kyunghyun Cho, and Jiatao Gu. 2020.
\newblock \href {https://doi.org/10.1609/aaai.v34i05.6451} {Neural machine
  translation with byte-level subwords}.
\newblock \emph{Proceedings of the AAAI Conference on Artificial Intelligence},
  34(05):9154--9160.

\bibitem[{Wei et~al.(2021)Wei, Liu, Guo, and Jiang}]{wei-2021-training}
Junqiu Wei, Qun Liu, Yinpeng Guo, and Xin Jiang. 2021.
\newblock \href {http://arxiv.org/abs/2101.09469} {Training multilingual
  pre-trained language model with byte-level subwords}.
\newblock \emph{CoRR}, abs/2101.09469v2.

\bibitem[{Wies et~al.(2021)Wies, Levine, Jannai, and
  Shashua}]{wies-2021-transformer}
Noam Wies, Yoav Levine, Daniel Jannai, and Amnon Shashua. 2021.
\newblock \href {https://proceedings.mlr.press/v139/wies21a.html} {Which
  transformer architecture fits my data? a vocabulary bottleneck in
  self-attention}.
\newblock In \emph{Proceedings of the 38th International Conference on Machine
  Learning}, volume 139 of \emph{Proceedings of Machine Learning Research},
  pages 11170--11181. PMLR.

\bibitem[{Wu et~al.(2016)Wu, Schuster, Chen, Le, Norouzi, Macherey, Krikun,
  Cao, Gao, Macherey, Klingner, Shah, Johnson, Liu, Kaiser, Gouws, Kato, Kudo,
  Kazawa, Stevens, Kurian, Patil, Wang, Young, Smith, Riesa, Rudnick, Vinyals,
  Corrado, Hughes, and Dean}]{wu-2016-gnmt}
Yonghui Wu, Mike Schuster, Zhifeng Chen, Quoc~V. Le, Mohammad Norouzi, Wolfgang
  Macherey, Maxim Krikun, Yuan Cao, Qin Gao, Klaus Macherey, Jeff Klingner,
  Apurva Shah, Melvin Johnson, Xiaobing Liu, Lukasz Kaiser, Stephan Gouws,
  Yoshikiyo Kato, Taku Kudo, Hideto Kazawa, Keith Stevens, George Kurian,
  Nishant Patil, Wei Wang, Cliff Young, Jason Smith, Jason Riesa, Alex Rudnick,
  Oriol Vinyals, Greg Corrado, Macduff Hughes, and Jeffrey Dean. 2016.
\newblock \href {http://arxiv.org/abs/1609.08144} {Google's neural machine
  translation system: Bridging the gap between human and machine translation}.
\newblock \emph{CoRR}, abs/1609.08144v2.

\bibitem[{Xiong et~al.(2019)Xiong, Wu, Wang, Kulkarni, Yu, Chang, Guo, and
  Wang}]{xiong-etal-2019-tweetqa}
Wenhan Xiong, Jiawei Wu, Hong Wang, Vivek Kulkarni, Mo~Yu, Shiyu Chang,
  Xiaoxiao Guo, and William~Yang Wang. 2019.
\newblock \href {https://doi.org/10.18653/v1/P19-1496} {{TWEETQA}: A social
  media focused question answering dataset}.
\newblock In \emph{Proceedings of the 57th Annual Meeting of the Association
  for Computational Linguistics}, pages 5020--5031, Florence, Italy.
  Association for Computational Linguistics.

\bibitem[{Xue et~al.(2021)Xue, Constant, Roberts, Kale, Al-Rfou, Siddhant,
  Barua, and Raffel}]{mt5}
Linting Xue, Noah Constant, Adam Roberts, Mihir Kale, Rami Al-Rfou, Aditya
  Siddhant, Aditya Barua, and Colin Raffel. 2021.
\newblock \href {https://doi.org/10.18653/v1/2021.naacl-main.41} {m{T}5: A
  massively multilingual pre-trained text-to-text transformer}.
\newblock In \emph{Proceedings of the 2021 Conference of the North American
  Chapter of the Association for Computational Linguistics: Human Language
  Technologies}, pages 483--498, Online. Association for Computational
  Linguistics.

\bibitem[{Yang et~al.(2019)Yang, Zhang, Tar, and
  Baldridge}]{yang-etal-2019-paws}
Yinfei Yang, Yuan Zhang, Chris Tar, and Jason Baldridge. 2019.
\newblock \href {https://doi.org/10.18653/v1/D19-1382} {{PAWS}-{X}: A
  cross-lingual adversarial dataset for paraphrase identification}.
\newblock In \emph{Proceedings of the 2019 Conference on Empirical Methods in
  Natural Language Processing and the 9th International Joint Conference on
  Natural Language Processing (EMNLP-IJCNLP)}, pages 3687--3692, Hong Kong,
  China. Association for Computational Linguistics.

\bibitem[{Yu et~al.(2020)Yu, Vu, and Kuhn}]{yu-etal-2020-ensemble}
Xiang Yu, Ngoc~Thang Vu, and Jonas Kuhn. 2020.
\newblock \href {https://doi.org/10.18653/v1/2020.sigmorphon-1.5} {Ensemble
  self-training for low-resource languages: Grapheme-to-phoneme conversion and
  morphological inflection}.
\newblock In \emph{Proceedings of the 17th SIGMORPHON Workshop on Computational
  Research in Phonetics, Phonology, and Morphology}, pages 70--78, Online.
  Association for Computational Linguistics.

\bibitem[{Zhang et~al.(2020)Zhang, Zhao, Saleh, and Liu}]{zhang-2020-pegasus}
Jingqing Zhang, Yao Zhao, Mohammad Saleh, and Peter Liu. 2020.
\newblock \href {http://proceedings.mlr.press/v119/zhang20ae.html} {{PEGASUS}:
  Pre-training with extracted gap-sentences for abstractive summarization}.
\newblock In \emph{Proceedings of the 37th International Conference on Machine
  Learning}, volume 119 of \emph{Proceedings of Machine Learning Research},
  pages 11328--11339. PMLR.

\bibitem[{Zhang et~al.(2015)Zhang, Zhao, and LeCun}]{xiang-2015-character}
Xiang Zhang, Junbo Zhao, and Yann LeCun. 2015.
\newblock \href
  {https://proceedings.neurips.cc/paper/2015/file/250cf8b51c773f3f8dc8b4be867a9a02-Paper.pdf}
  {Character-level convolutional networks for text classification}.
\newblock In \emph{Advances in Neural Information Processing Systems},
  volume~28. Curran Associates, Inc.

\bibitem[{Zilly et~al.(2017)Zilly, Srivastava, Koutn\'{\i}k, and
  Schmidhuber}]{zilly-2017-recurrent-highway-nets}
Julian~Georg Zilly, Rupesh~Kumar Srivastava, Jan Koutn\'{\i}k, and J{\"u}rgen
  Schmidhuber. 2017.
\newblock \href {http://proceedings.mlr.press/v70/zilly17a.html} {Recurrent
  highway networks}.
\newblock In \emph{Proceedings of the 34th International Conference on Machine
  Learning}, volume~70 of \emph{Proceedings of Machine Learning Research},
  pages 4189--4198. PMLR.

\end{thebibliography}
\bibliographystyle{acl_natbib}

\end{document}